\documentclass{article}

\usepackage[nonatbib, main, final]{neurips_2026}
\usepackage[utf8]{inputenc} %
\usepackage[T1]{fontenc}    %

\usepackage{url}            %
\usepackage{booktabs}       %
\usepackage{amsfonts}       %
\usepackage{nicefrac}       %
\usepackage{microtype}      %
\usepackage{caption,makecell}
\usepackage{xcolor}         %
\usepackage{multirow}
\usepackage{amsmath,amsthm,amssymb,mathtools}        %
\usepackage{braket}
\usepackage{subcaption}
\usepackage{wrapfig}
\usepackage{graphicx}       %
\usepackage{siunitx}  %
\usepackage{textcomp}   %
\usepackage[inline]{enumitem}  

\usepackage{comment}
\usepackage[autostyle]{csquotes}
\usepackage[backend=biber, natbib=true, sorting=none]{biblatex}

\usepackage{xspace}

\usepackage{tikz}
\usetikzlibrary{shapes.geometric, arrows.meta}

\renewcommand{\vec}{\mathbf}

\newcommand{\second}[1]{\textcolor{green!40!black}{#1}} %

\newcommand{\netname}{Dandelion\xspace}

\newcommand{\galewskytt}{\texttt{barotropic-jets}\xspace}
\newcommand{\mickelintt}{\texttt{active-turbulence}\xspace}
\newcommand{\cahnhilliardtt}{\texttt{cahn-hilliard}\xspace}
\newcommand{\shockcapstt}{\texttt{shock-caps}\xspace}
\newcommand{\heldsuareztt}{\texttt{dry-atmosphere-3D}\xspace}
\newcommand{\oceantt}{\texttt{global-ocean-3D}\xspace}

\usepackage{hyperref}       %
\usepackage{cleveref}

\title{Dandelion: A Spherical Flower for Neural Simulation of Planetary Dynamics}

\author{%
  Till Muser \\
  University of Basel\\
  \texttt{till.muser@unibas.ch}
  \And
  Giovanni Abati\\
  University of Basel\\
  \texttt{giovanni.abati@unibas.ch}
  \And
  Ivan Dokmanić\\
  University of Basel\\
  \texttt{ivan.dokmanic@unibas.ch}
}

\begin{document}

\maketitle

\begin{abstract}
Many dynamical processes unfold on the sphere but the default scientific machine learning architectures are Euclidean. Applying these architectures on a regular lat--lon grid causes problems: Cartesian convolutions become distorted at high latitude; 2D FFTs in Fourier neural operators incorrectly assume double periodicity; Cartesian positional encodings in ViTs distort spherical geodesic distances. Recent work moves towards natively spherical  primitives, including spherical convolutions (e.g., DeepSphere or DISCO), Spherical Fourier Neural Operators (SFNOs), and geodesic attention. Here we propose \netname, a spherical version of Flower---a warp-based neural PDE solver. A \netname layer computes a tangent-plane displacement and then  transports features along great circles. We obtain a U-Net-like structure by implementing hierarchical pooling entirely in the spherical-harmonic domain. There are thus no convolutions: spatial mixing is  achieved only through spherical coordinate changes---or warps.
To compare \netname with existing spherical architectures, we release an evolving benchmark suite of challenging, natively spherical PDE datasets including a modified Galewsky jet, anomalous chained turbulence, Cahn--Hilliard decomposition, spherical Riemann shocks, Held--Suarez dry atmospheric transport and global ocean dynamics. This new benchmark fills the gap between existing spherical datasets which are either too small and stylized, or much too large (ERA5) for model iteration.
\netname achieves the best performance on all but one benchmark setting, where the best model is Flower2D, also based on warps. The spherical primitives result in the most physically consistent forecasts.
\end{abstract}

\section{Introduction}
\label{sec:introduction}

In a number of applications, neural PDE solvers have become a competitive substitute for classical numerical solvers. A flagship success of this paradigm is global weather forecasting~\cite{pathak2022fourcastnetglobaldatadrivenhighresolution, bi2022panguweather3dhighresolutionmodel, bodnar2024aurora, lam2023graphcast} where deep-learning  surrogates can match the skill of large-scale numerical codes while running orders of magnitude faster. Similar ideas are now being applied to oceanography~\cite{yuan2023space, rajagopal2023evaluation,  choi2024applications, xu2025prediction}, atmospheric dynamics on other planets~\cite{schmude2026pdefoundationmodelsskillful}, or other parts of astrophysics~\cite{branca2024emulating, van2025bridging, al2026fourier}.

These planetary dynamics are supported on the sphere $S^2 \subset \mathbb{R}^3$, whereas the standard architectural toolkit assumes flat Euclidean grids. Cartesian convolutions, vision-transformer tokenizers, and grid-index positional encodings are topologically and geometrically mismatched with the sphere. This deteriorates the inductive biases of these architectures resulting in poor sample complexity and sensitivity to rotations. A high-capacity model can memorize latitudinal forcing patterns to fit the data instead of learning the fluid physics, which would generalize across rotations.

Several recent architectures address this problem. Spectral approaches such as the Spherical Fourier Neural Operator~\cite{bonev2023sphericalfourierneuraloperators} parametrize  \emph{global} spatial mixing in the spherical-harmonic domain, preserving continuous $\mathrm{SO}(3)$ symmetry by construction; the same recipe scales up to weather-grade systems in SFNO's descendant FourCastNet~3~\cite{bonev2025fourcastnet3geometricapproach}. Equivariant and geometric-mesh approaches replace the lat--lon grid altogether: spherical CNNs parametrize $\mathrm{SO}(3)$-equivariant convolutions on the rotation group~\cite{cohen2018sphericalcnns, ocampo2023scalableequivariantsphericalcnns}, graph models like GraphCast~\cite{lam2023graphcast} implement message passing on icosahedral meshes, and HEALPix~\cite{gorski2005healpix} supplies an equal-area discretization widely used in astrophysics. \citet{bonev2025attentionsphere} recently adapted localized attention to $S^2$, using numerical quadrature weights and geodesic neighborhood windows to keep local windowed attention geometrically faithful.

We propose a new direction based on the recently proposed Flower architecture~\cite{muser2026flowerswarpdriveneural}, in which each layer is a (multihead) learned warp (a coordinate change), that predicts a per-pixel displacement field and reads its features at the displaced positions, at cost linear in the grid size. \netname is built on the same principle, but it is designed to be  natively spherical. Per-pixel displacements live in the tangent space and the warp transports features along great circles (spherical geodesics), rather than along straight lines in pixel space. Strided pooling is replaced by spectral coarsening, which results in efficient computation, similarly as in the Euclidean Flower that we draw inspiration from (\S\ref{sec:methodology}).

\begin{figure}[t]
    \centering
    \includegraphics[width=\linewidth]{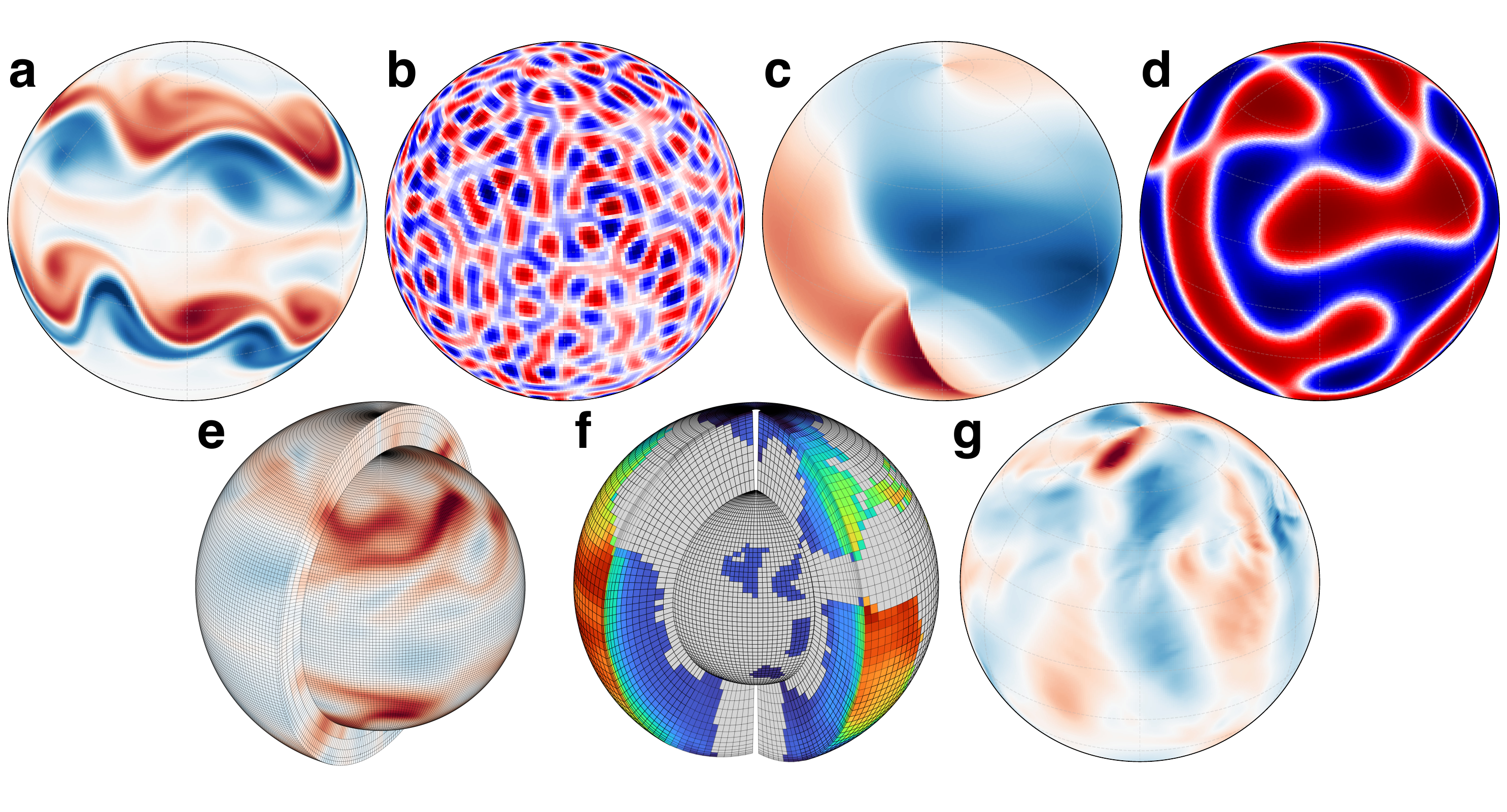} %
    \caption{Snapshots of the seven datasets: \textbf{(a)} \galewskytt (vorticity), \textbf{(b)} \mickelintt (vorticity), \textbf{(c)} \shockcapstt (momentum $u$), \textbf{(d)} \cahnhilliardtt ($\phi_0$), \textbf{(e)} \heldsuareztt (zonal wind $u$ at various pressure levels), \textbf{(f)} \oceantt (temperature $T$ at depth levels), and \textbf{(g)} \texttt{planetswe} (azimuthal velocity). \textbf{(a)}--\textbf{(f)} are introduced by this paper; \textbf{(g)} is from The Well~\cite{ohana2025welllargescalecollectiondiverse}.}
    \label{fig:eight-snapshots}
\end{figure}

We complement the proposed architecture with a new benchmark suite that fills a gap in the spherical PDE dataset landscape. Euclidean architectures can be evaluated on many benchmarks~\cite{takamoto2024pdebenchextensivebenchmarkscientific, tali2024flowbenchlargescalebenchmark, ohana2025welllargescalecollectiondiverse}, but no comparable suite exists for the sphere: spherical architectures are typically tested either on a handful of toy problems---forced 2D shallow water without topography being a canonical example~\cite{bonev2023sphericalfourierneuraloperators, bonev2025attentionsphere}---or, at the other extreme, on supermassive operational reanalyses such as ERA5 through WeatherBench~2~\cite{rasp2024weatherbench2}. Both extremes are unsatisfactory for method development. Toy problems do not enable a thorough comparison between methods; ERA5-scale training, by contrast, makes architectural debugging and ablation studies prohibitively expensive, with a single hyperparameter sweep easily running into thousands of GPU-days. We thus designed our benchmark to occupy the middle ground: physically diverse enough to stress-test architectural choices, but small enough to train from scratch in hours rather than weeks.

We contribute six new challenging, physically diverse spherical PDE datasets (see \autoref{tab:datasets}). Together with the \texttt{planetswe} dataset from The Well~\cite{ohana2025welllargescalecollectiondiverse} this yields a seven-dataset suite (\S\ref{sec:datasets}, \autoref{fig:eight-snapshots}). %
The phenomena span barotropic instability, active-matter turbulence, phase separation, hyperbolic shocks, dry GCM dynamics, and global ocean dynamics. Together they form a substantially broader physical envelope than any single existing spherical benchmark. At the same time, datasets are small or moderate in size so that a model can be trained from scratch in hours rather than weeks. This enables continuous testing and debugging of candidate spherical architectures before committing to operational data. We use this new benchmark to evaluate \netname in \S\ref{sec:experiments} against SFNO, FNO, the local $S^2$/$R^2$ transformer baselines from \citet{bonev2025attentionsphere}, and the original 2D Flower. All datasets and our code are publicly available.\footnote{Datasets: \url{https://huggingface.co/collections/sada-group/spherical-pde-datasets}; code: \url{https://github.com/t-muser/dandelion}.}

\section{Methodology: \netname}
\label{sec:methodology}

The Flower architecture~\cite{muser2026flowerswarpdriveneural} is a U-Net whose only spatial-mixing primitive is a multi-head warp.
Each block contains $H$ parallel heads.
A single pointwise value map $V$ is applied to the block input $u$, and the resulting
field $v = Vu$ is split channelwise across heads as in multi-head attention,
$v = v^{(1)} \oplus \cdots \oplus v^{(H)}$. In parallel, a pointwise map $g$ predicts
one displacement field per head, $\delta^{(h)}(x) = g^{(h)}(u(x))$. Head $h$ then
samples its value slice at the displaced location,
\begin{equation}
    u_\text{out}^{(h)}(x) = v^{(h)}\!\left(x + \delta^{(h)}(x)\right),
\end{equation}
evaluated via bilinear \texttt{grid\_sample}; per-head outputs are concatenated and combined with a residual. The construction follows the flow-map view of conservation laws: instead of aggregating a neighborhood through a fixed kernel or a spectral multiplier, each head pulls a feature from a single, input-dependent source location ``upstream''.
Stacked at every U-Net level, these warps act as a learned transport operator with cost linear in the grid size, and the architecture contains no Fourier mixing, no attention, and no convolutional spatial mixing beyond the strided pooling between levels. Despite this minimalism, Flower is highly competitive: at 17M parameters it outperforms similarly-sized spectral, convolutional, and attention
baselines across sixteen 2D and 3D PDE benchmarks from The Well and PDEBench, and a
150M-parameter variant trained from scratch surpasses the pretrained 628M-parameter Poseidon-L foundation model~\cite{herde2024poseidon} on compressible Euler.

However, several of Flower's primitives assume a flat geometry. A pixel-space displacement corresponds to a latitude-dependent arc length, since equiangular cells near the poles are much smaller than at the equator. Periodic padding is correct in longitude but not in latitude; applying it in both directions treats the grid as a torus, whereas on the sphere a row near the pole connects across the pole to the row at the opposite longitude. The square receptive field of strided $2 \times 2$ pooling has a physical extent that varies with latitude, giving the multi-resolution hierarchy uneven coverage. These distortions are mild near the equator but grow towards the poles. 
\netname keeps the warp-based U-Net and replaces each grid-sensitive primitive with a native $S^2$ analogue; we expect these primitives to matter most when dynamics reach high latitudes or cross the poles.

\begin{figure}[t]
    \centering
    \begin{subfigure}[b]{0.31\textwidth}
        \centering
        \includegraphics[width=\linewidth]{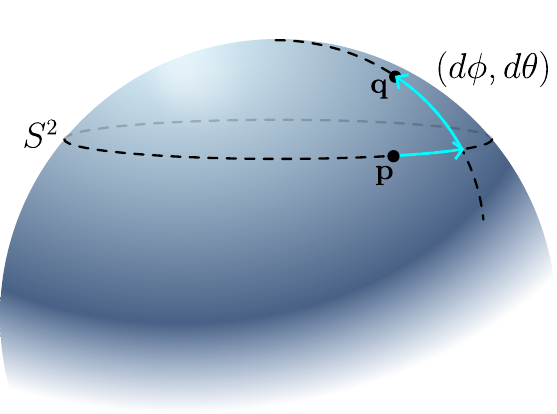}
        \caption{Lat-lon displacement}
        \label{fig:latlondisp}
    \end{subfigure}
    \hspace{0.08\textwidth}
    \begin{subfigure}[b]{0.31\textwidth}
        \centering
        \includegraphics[width=\linewidth]{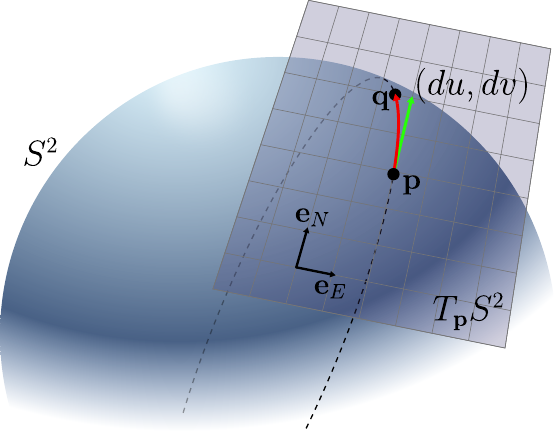}
        \caption{Tangent warp}
        \label{fig:tangentwarp}
    \end{subfigure}
    \caption{Two schemes for defining a sampling displacement on the sphere $S^2$.
    (\subref{fig:latlondisp}) Lat-lon displacement: an angular offset $(d\phi, d\theta)$ (light blue arrow) is added to the chart coordinates of the query point $\vec{p}$ to obtain the new sampling location $\vec{q}$.
    (\subref{fig:tangentwarp}) Tangent warp: a displacement $(du, dv)$ (green arrow) is predicted in the local tangent plane $T_\vec{p}S^2$ relative to the orthonormal basis $(\vec{e}_N, \vec{e}_E)$, then mapped to a point $\vec{q} \in S^2$ via the exponential map (Rodrigues' formula), which traces the great circle through $\vec{p}$ in the predicted tangent direction
    (red arrow).}
    \label{fig:architecture}
\end{figure}

\paragraph{Tangent warping} We now introduce basic notions from differential geometry on which we build \netname. More details can be found in standard textbooks such as~\cite{lee2018introduction}. We write $T_p M$ for the tangent space at a point $p$ on a Riemannian manifold $(M, g)$; see~\autoref{fig:tangentwarp}. The metric $g$ induces, at each $p \in M$, an inner product $g_p \colon T_p M \times T_p M \to \mathbb{R}$ on its tangent space.
This lets us assign lengths to smooth curves
$\gamma \colon [a, b] \to M$ and define geodesics as locally length-minimizing curves between two points. Then, at each point $p \in M$, one defines the \emph{exponential map} $\exp_p \colon T_p M \to M$ as the map sending an initial velocity vector $\vec{v} \in T_p M$ to the endpoint of the unique geodesic $\gamma_\vec{v} \colon [0, 1] \to M$ satisfying $\gamma_\vec{v}(0) = p$ and $\dot{\gamma}_\vec{v}(0) = \vec{v}$.
For \netname, the manifold is the unit sphere $S^2 \subset \mathbb{R}^3$, and points on it are in one-to-one correspondence with unit vectors $\vec{p} \in \mathbb{R}^3$, $\| \vec{p} \| = 1$. With the Euclidean inner product on tangent planes, the exponential map is given in closed form by Rodrigues's formula, i.e., for a tangent vector $\vec{v} \in T_\vec{p}S^2$ one obtains $\exp_\vec{p}(\vec{v})$ by rotating $\vec{p}$ around the axis $\vec{p} \times \vec{v}$ by an angle $\lVert\vec{v}\rVert$,
\begin{equation}
    \exp_\vec{p}(\vec{v}) = \vec{p} \cos ( \lVert \vec{v} \rVert) + \frac{\vec{v}}{\lVert\vec{v}\rVert} \sin(\lVert \vec{v}\rVert).
\end{equation}
Since the plane spanned by $\vec{p}$ and $\vec{v}$ intersects the sphere along a great circle, and Rodrigues' formula describes a rotation of $\vec{p}$ in that plane, geodesics on a sphere are great circles.

This construction addresses the fact that displacements on embedded manifolds cannot be treated as ordinary translations. While for $M= \mathbb{R}^n$ every tangent space $T_p\mathbb{R}^n$ is canonically identified with the manifold itself, $T_p\mathbb{R}^n \cong \mathbb{R}^n$, this is in general not true for arbitrary manifolds: points on the sphere live in $S^2$, but tangent vectors in $T_p S^2$. Although both can be represented in the ambient space $\mathbb{R}^3$, the ambient sum $\vec{p} + \vec{v}$ generally leaves the sphere, and does not define an intrinsic displacement on $S^2$.  On a global level, this is reflected on the fact that no flat coordinate system can faithfully represent the spherical metric everywhere (more informally, there are no perfectly faithful world maps). Planar parameterizations, such as lat-lon grids, necessarily introduce distortions, singularities or cuts. This is why Euclidean neural architectures are suboptimal for spherical data.

We now define a warp for spherical data: given a query point $\vec{p} \in S^2$, the model should predict a displacement determining the next sampling point $\vec{q}$. A naive approach would be to apply the displacement directly in the coordinates in which the data are stored; for a latitude-longitude grid, this would amount to predicting angular offsets $(d\phi, d\theta)$, as depicted in~\autoref{fig:latlondisp}. Combined with suitable boundary conditions, this is a plausible coordinate-based baseline. But fixed angular displacements do not correspond to fixed geodesic distances on $S^2$, so their metric is latitude-dependent.

We use a more principled approach and define displacement intrinsically using the introduced geometric machinery: \netname represents a query point $\vec{p} \in S^2$ as a unit vector in $\mathbb{R}^3$, and instead of operating in the pixel space it predicts a displacement $du \vec{e}_N + dv \vec{e}_E \in T_\vec{p} S^2$, where $\vec{e}_N$ and $\vec{e}_E$ denote the northward and eastward directions (see~\autoref{fig:tangentwarp}). This is mapped back to $\vec{q} \in S^2$ through Rodrigues's formula, rotating $\vec{p}$ along the great circle identified by the predicted tangent direction. Since on $S^2$ the norm of the tangent vector $\lVert \vec{v} \rVert$ equals the arc length traveled along the geodesic, for a given predicted $\vec{v}$ the warp distance is latitude-invariant, yielding a receptive field that is independent of distortions introduced by the sphere’s various Euclidean parameterizations.

\paragraph{Spectral coarsening} The second native primitive we need is hierarchical pooling. The Flower paper showed that arranging warping blocks in a multiscale U-Net-like structure gives the best performance. This means that we have to process the input at a sequence of resolutions, alternating coarsening steps that produce a low-resolution summary with refinement steps that recover the fine grid, and joining the two paths by skip connections. Each coarsening step must  retain coarse-scale content while discarding the fine scales that would be aliased on the coarser grid. Indeed, naive downsampling without a  low-pass filter would result in aliasing~\cite{bartolucci2023representation, chaman2021trulyshiftinvariantconvolutionalneural}. We thus proceed in two stages: project onto a band-limited subspace, then evaluate on the coarser grid. The first stage carries all the geometric content; the second discards a subset of grid points.

To this end we use spherical harmonics~\cite{vilenkin1978special} $(Y_\ell^m, \ell \geq 0, |m| \leq \ell)$, which is the standard Fourier basis on $S^2$ which diagonalizes the corresponding Laplace--Beltrami operator.
Any $f \in L^2(S^2)$ admits the harmonic expansion $f = \sum_{\ell=0}^{\infty}\sum_{m=-\ell}^{\ell} \hat{f}_\ell^m Y_\ell^m$. Band-limiting at degree $L$ (lowpass filtering) truncates this expansion,
\begin{equation}
    \mathbb{P}_L f =
    \sum_{\ell=0}^{L}\sum_{m=-\ell}^{\ell}
    \hat{f}_\ell^m\, Y_\ell^m,
    \quad
    \hat{f}_\ell^m =
    \int_{S^2} f(\mathbf{p})\, Y_\ell^m(\mathbf{p})\, \mathrm{d}\Omega(\mathbf{p}).
\end{equation}
We refer the reader to standard references like~\cite{vilenkin1978special} for more details.

$\mathbb{P}_L$ is a natural primitive for hierarchical pooling on $S^2$. First, the construction is fully intrinsic: it depends only on the spectrum of $\Delta_{S^2}$ and never references a coordinate chart, mirroring the desideratum for tangent warping. Second, each $\mathcal{H}_\ell$ is an irreducible representation of the rotation group $\mathrm{SO}(3)$. A rotation $R \in \mathrm{SO}(3)$ acts on functions by pullback, $f \mapsto f \circ R^{-1}$; $\mathcal{H}_\ell$ are its invariant subspaces so $\mathbb{P}_L$ commutes with the rotation action. That is, $\mathbb{P}_L(f \circ R^{-1}) = (\mathbb{P}_L f)\circ R^{-1}$ for any $R \in \mathrm{SO}(3)$, and resolution change preserves the rotational symmetry of the physics by construction.

A strided convolution on the lat--lon grid satisfies neither property. The spherical area of an equiangular cell at colatitude $\theta$ scales as $\sin\theta$, so a fixed-size pooling stencil aggregates physical regions whose areas vary by orders of magnitude between equator and poles. The operation is also not rotation-equivariant, so any model built on it can only recover rotational symmetry as an empirical regularity, never as a structural invariant.

\netname realizes $\mathbb{P}_L$ at every resolution change through a forward Spherical Harmonic Transform (SHT) onto the truncated coefficient set $\{\hat f_\ell^m : \ell \leq L\}$, followed by an inverse SHT onto the target grid; a pointwise linear mixing adjusts the channel count and is the only convolution-shaped operator left in the spatial-mixing path. The spatial mixing  happens entirely through the warp. Down- and up-sampling differ only in the choice of target grid: the downsampler evaluates the inverse transform on the coarse grid with $L$ matched to its Nyquist degree, while the upsampler evaluates the same truncated coefficients on the fine grid. The continuous round-trip is, by construction, a low-pass projection. We use the SHT implementation of \texttt{torch-harmonics}~\cite{bonev2023sphericalfourierneuraloperators}.

\section{Datasets}
\label{sec:datasets}

We now describe the datasets we use for benchmarking. More details are given in~\autoref{app:dataset}.

\paragraph{Earth-like Shallow Water (\texttt{planetswe})}
A reference dataset from The Well~\cite{ohana2025welllargescalecollectiondiverse, mccabe2023towards}, inspired by Problem 7 of Williamson's classic test suite~\cite{williamson1992testset}. \texttt{planetswe} solves forced, hyperviscous, rotating shallow water on a sphere with Earth-like topography and periodic daily and annual forcings, initialised from ERA5 at the $500\,$hPa level. The continuous forcing means the dynamics never decay, so that the benchmark addresses long-horizon stability; the diurnal cycle is locked to the rotation axis, which forces a model to either be time-aware or to pick up the periodicity from context.
See the \href{https://polymathic-ai.org/the_well/datasets/planetswe}{data sheet of the \texttt{planetswe}} on The Well's project page for a visualization  and additional information.

\paragraph{Rotated Double Galewsky (\galewskytt)}
This dataset is related to \texttt{planetswe} in the same way that Galewsky's original test~\cite{galewsky:paper} is related to Williamson Problem~7: where \texttt{planetswe} stresses long-horizon stability under realistic forcing and fixed geometry, \galewskytt isolates rotational and geometric reasoning in a freely-evolving, instability-driven regime. The base test features a single compact-support zonal jet with a %
perturbation that triggers barotropic instability, which we extend along two axes. We add a mirrored stable-state jet in the opposite hemisphere to probe cross-equatorial wave propagation, which is absent with a single jet. We also randomly rotate the entire configuration to remove any grid-aligned reference frame, which forces the model to infer the Coriolis parameter dynamically from the velocity field rather than memorize it from a fixed axis (as it can in \texttt{planetswe}). See \autoref{fig:galewsky-4-evolution} for a visualization of how a trajectory evolves.

\paragraph{Anomalous Chained Turbulence (\mickelintt)} Based on the covariant generalized Navier-Stokes model for actively driven fluids~\cite{mickelin:paper}, this dataset simulates non-equilibrium flows on a sphere. The system shows three different regimes: a quasistationary burst phase ($B$ phase, with too few unstable modes to develop turbulence, and excluded from our splits), a classic Kolmogorov 2D turbulence ($T$ phase), and, most notably, an anomalous turbulence, with finite-size vortices self-organizing into percolating, antiferromagnetic-ordered chains ($A$ phase).
This antiferromagnetic chain order is a closed-manifold effect that has no flat analogue, making the $A$-phase uniquely spherical.
See \autoref{fig:mickelin-4-samples} for a visualization of the corners of the parameter space.

\paragraph{Spinodal Decomposition (\cahnhilliardtt)} The Cahn-Hilliard equation describes the phase separation of binary fluids into component-pure domains~\cite{cahnhilliard:paper}, and serves as a rigorous test for high-order spatial derivatives and conservation laws. In particular, the presence of the $\nabla^4$ biharmonic operator amplifies the sensitivity to high-frequency modes and imposes strict smoothness constraints. Total mass conservation further requires that the total composition remain constant over time. This is a test of whether learned models can preserve global invariants without introducing drifts.
See \autoref{fig:cahn-hilliad-4-samples} for a visualization of the corners of the parameter space.

\begin{table}[t]
    \centering
    \small
    \caption{Summary of the provided spherical PDEs benchmark datasets. Sizes are the compressed HDF5 releases; dataset names link to their HuggingFace repositories.}
    \begin{tabular}{llS[table-format=3.0]S[table-format=3.0]}
    \toprule
    Dataset & Description & {Runs} & {Size (GB)}\\
    \midrule
    \href{https://huggingface.co/datasets/sada-group/barotropic-jets}{\galewskytt}
        & Barotropic instability from shallow water jets
        & 960
        & 210 \\
    \href{https://huggingface.co/datasets/sada-group/active-turbulence}{\mickelintt}
        & Anomalous chained antiferromagnetic turbulence
        & 760
        & 18 \\
    \href{https://huggingface.co/datasets/sada-group/cahn-hilliard}{\cahnhilliardtt}
        & Spinodal decomposition
        & 576
        & 15 \\
    \href{https://huggingface.co/datasets/sada-group/shock-caps}{\shockcapstt}
        & Multi-shock shallow water Riemann problem
        & 500
        & 63 \\
    \href{https://huggingface.co/datasets/sada-group/dry-atmosphere-3D}{\heldsuareztt}
        & 3D Held-Suarez global atmospheric simulation 
        & 162
        & 667 \\
    \href{https://huggingface.co/datasets/sada-group/global-ocean-3D}{\oceantt}
        & 3D global oceanic simulation
        & 243
        & 248 \\
    \bottomrule
    \end{tabular}
    \label{tab:datasets}
\end{table}

\paragraph{Shallow Water Shocks (\shockcapstt)} A spherical adaptation of the 2D Riemann problem. Each trajectory starts from $K$ geodesic caps placed uniformly at random on the sphere, each carrying its own piecewise-constant state $(h, u, v)$ over an equally random background. Evolving under the shallow-water equations, the discontinuities in the depth $h$ at cap boundaries are sufficient on their own to drive Riemann fans, so the dataset stays interesting even at zero velocity. We vary two parameters across runs: the cap count $K$ and the velocity scale $\delta$. As the fronts propagate they collide, produce expanding bores, and eventually focus at the antipodes of their sources.
See \autoref{fig:shock-caps-4-evolution} for a visualization of how a trajectory evolves.
\paragraph{Held--Suarez Dry Atmospheric Transport (\heldsuareztt)} The Held--Suarez~\cite{held1994proposal} system is a standard example of an atmospheric Global Circulation Model (GCM). It features two simple forcing and dissipation dynamics: a Newtonian temperature relaxation to a zonally-symmetric equilibrium state, and a Rayleigh damping on the surface representing boundary-layer friction.
We vary the rotational speed (angular velocity) $\Omega$, the equator-to-pole temperature contrast $\Delta T_y$, and the static stability $\Delta\theta_z$. The trajectories cover a parameter space across which number, latitude, and width of the eddy-driven jets vary.
See \autoref{fig:held-suarez-4-evolution} for a visualization of how a trajectory evolves.
\paragraph{Ocean Dynamics (\oceantt)} A 3D, $100$-year global oceanic simulation with realistic bathymetry, tracking potential temperature, salinity, velocity, and sea level~\cite{marshall1997finite}. The presence of seasonality patterns poses an additional challenge for learned models, since it tests their capability to forecast non-autonomous dynamical systems with complex boundary conditions. \autoref{fig:ocean-dynamics-4-seasons} shows trajectory snapshots of different seasons. 

\section{Results}
\label{sec:experiments}

We benchmark \netname against six baselines on the seven datasets of \S\ref{sec:datasets}: the Euclidean Flower2D~\cite{muser2026flowerswarpdriveneural}, FNO~\cite{li2021fourierneuraloperatorparametric}, SFNO~\cite{bonev2023sphericalfourierneuraloperators}, and the local $R^2$ and $S^2$ neighborhood transformers of~\citet{bonev2025attentionsphere}. This covers three primitives (Fourier layer, attention, warping), each in a Euclidean and a spherical variant. Additionally, we compare against GAOT~\cite{wen2026geometryawareoperatortransformer}, which encodes point coordinates to a latent grid, applies a Vision transformer there, and then decodes back.

All models share the same $4\!\to\!1$ next-step task and the same two-phase training schedule: $20$ epochs of single-step supervision followed by $5$ epochs of two-step rollout, with gradients flowing through both unrolls. This is the recipe used to train data-driven weather models such as GraphCast~\cite{lam2023graphcast}, though scaled down; those works typically extend the rollout phase to many more steps. For this benchmark we treat the 3D datasets as 2D fields stacked along the channel dimension, even though they are fully 3D; \S\ref{sec:2p5d} evaluates a 2.5D variant of \netname that keeps the vertical structure. The learning rate is swept over $\{1\times10^{-4},\, 5\times 10^{-4},\, 1\times10^{-3}\}$ and the per-architecture parameter count for FNO, SFNO, Flower2D and Dandelion matched at 20M as closely as each model family allows. GAOT and the two neighborhood transformers were run with the hyperparameters from their respective publications, giving $\sim$3.4M parameters for GAOT and $\sim$500K for the transformers. We initially scaled every model to $\sim$20M parameters, but at that size the transformers sometimes needed several days per run. %
Full training and evaluation details are in \S\ref{app:experiment}. \autoref{tab:results41} reports area-weighted VRMSE for single-step prediction and 20-step autoregressive rollout.

\begin{table*}[b]
  \centering
  \small
  \setlength{\tabcolsep}{4pt}
    \caption{Area-weighted VRMSE for unconditioned 4$\to$1 next-step prediction and 1:20 rollout. Best in bold, second best colored green. See \autoref{eq:vrmse} for a definition of the metric. FNO, SFNO, Flower2D and \netname are matched at $\sim$20M parameters; GAOT ($\sim$3.4M) and the two neighborhood transformers ($\sim$500K) use the configurations published by their authors.}
  \begin{tabular}{@{}ll ccccccc@{}}
    \toprule
    & Dataset
      & \textsc{FNO}
      & \textsc{SFNO}
      & \textsc{GAOT}
      & \textsc{LR$^2$Trnsf.}
      & \textsc{LS$^2$Trnsf.}
      & \textsc{Flower2D}
      & \textsc{\netname} \\
    \midrule
    \multirow{7}{*}{\rotatebox[origin=c]{90}{\textit{Next-step}}}
      & \texttt{planetswe} & 0.0125 & 0.0084 & 0.0383 & 0.0054 & 0.0039 & \second{0.0008} & \textbf{0.0005} \\
      & \galewskytt        & 0.0600 & 0.0209 & 0.0421 & 0.0148 & 0.0117 & \second{0.0055} & \textbf{0.0037} \\
      & \mickelintt        & 0.5816 & 0.4132 & 0.3172 & 0.2774 & 0.2052 & \second{0.0700} & \textbf{0.0441} \\
      & \cahnhilliardtt    & 1.2052 & 0.2933 & 0.4153 & 0.1373 & 0.0997 & \second{0.0426} & \textbf{0.0293} \\
      & \shockcapstt       & 0.2695 & 0.0814 & 0.1659 & 0.0431 & 0.0235 & \second{0.0160} & \textbf{0.0082} \\
      & \heldsuareztt      & 0.1704 & 0.1040 & 0.1018 & 0.0445 & 0.0408 & \second{0.0290} & \textbf{0.0274} \\
      & \oceantt           & 0.2187 & 0.1733 & 0.0923 & 0.0978 & 0.1160 & \second{0.0253}          & \textbf{0.0252} \\
    \midrule
    \multirow{7}{*}{\rotatebox[origin=c]{90}{\textit{1:20 Rollout}}}
      & \texttt{planetswe} & 0.2027 & 0.0866 & 0.0908 & 0.0755 & 0.5587 & \second{0.0171} & \textbf{0.0108} \\
      & \galewskytt        & 1.8770 & 0.3213 & 0.3356 & 0.6558 & 0.2123 & \second{0.0588} & \textbf{0.0422} \\
      & \mickelintt        & 1.0657 & 1.1637 & \second{0.9580} & 1.1709 & 0.9885 & 0.9583          & \textbf{0.8915} \\
      & \cahnhilliardtt    & 25.547 & 41.447 & \second{14.7524} & 28.374 & 26.345 & \textbf{7.2393} & 16.093 \\
      & \shockcapstt       & 3.3886 & 0.7873 & 0.3908 & 0.4715 & 0.1581 & \second{0.0635} & \textbf{0.0414} \\
      & \heldsuareztt      & 3.3965 & 0.6426 & 0.4324 & 0.5599 & 0.3717 & \second{0.2214} & \textbf{0.2067} \\
      & \oceantt           & 1.603 & 1.885 & 0.788 & 0.958 & 1.511 & \second{0.232} & \textbf{0.197} \\
    \bottomrule
  \end{tabular}
  \label{tab:results41}
\end{table*}

\netname is strictly best on all $7$ next-step cells and on $6/7$ rollout cells. %
Euclidean Flower2D is second on most datasets, suggesting that considerable gain comes already from the multiscale warp-based paradigm; the dedicated spherical primitives, however, bring about clear improvement, cutting Flower2D's rollout error by  $1.4\times$ on \galewskytt, $1.6\times$ on \texttt{planetswe}, and $1.5\times$ on \shockcapstt. The gain is smallest on \heldsuareztt, whose jets are zonally aligned and stay away from the poles. FNO is typically the weakest baseline; SFNO sits between FNO and the warp-based models, but trails the latter by a substantial margin.

The $\sim$500K-parameter neighborhood transformers are the strongest non-warp baselines on next-step prediction: on six of the seven datasets both LR$^2$Trnsf.\ and LS$^2$Trnsf.\ improve on FNO, SFNO, and GAOT, with the spherical variant ahead of its Euclidean counterpart; \oceantt is the exception on both counts. They nonetheless trail both warp-based models in every cell. Their advantage is less consistent under rollout: LS$^2$Trnsf.\ is the best non-warp model on \galewskytt, \shockcapstt, and \heldsuareztt, but falls behind GAOT on \mickelintt, \cahnhilliardtt, and \oceantt.

\subsection{Rotation Equivariance on \galewskytt}
\label{sec:jets-equivariance}
\begin{figure}[t]
    \centering
    \includegraphics[width=\linewidth]{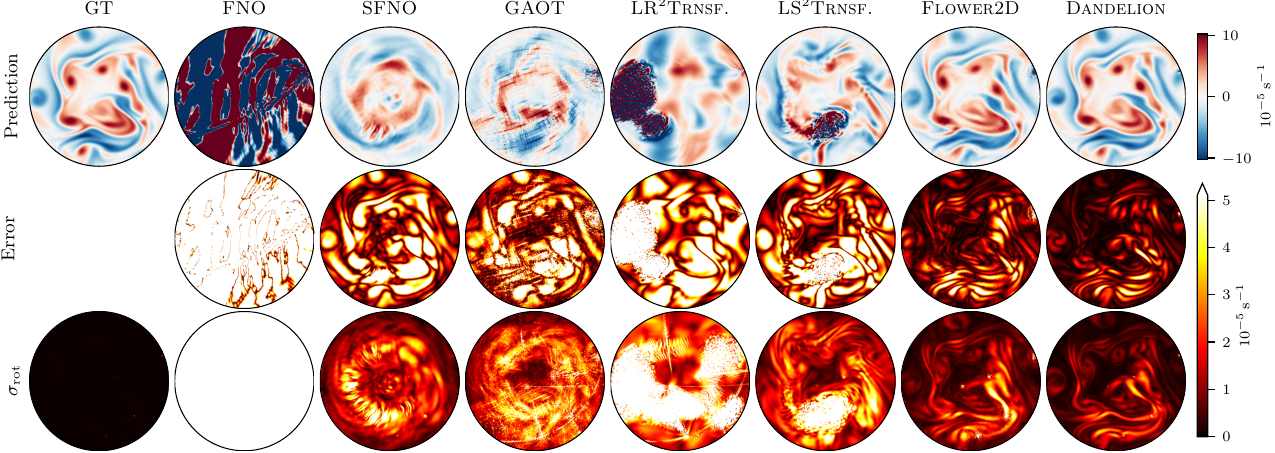}
    \caption{Vorticity $\zeta$ after a 20-step autoregressive rollout on one test configuration of \galewskytt. First row: prediction on one trajectory. Second row: absolute error of that prediction. Third row: equivariance test, pointwise standard deviation of the predictions over all six versions of this trajectory after realignment to the canonical frame (the quantity averaged in $\sigma_\text{rot}$, \autoref{eq:tilt-sigma}).}
    \label{fig:jets-pred}
\end{figure}

The \galewskytt dataset features six randomly rotated versions for each run (see~\autoref{app:datasets-galewsky}). It is then straightforward to measure how rotationally equivariant a trained model is: for each initial rotation, we roll out the model and rotate the forecast back to the canonical polar-aligned frame, $c_k(x) = R_k^{-1} f(R_k x)$, where $f$ is the forecast for tilt $k$; an exactly equivariant model would give $c_1 = \dots = c_6$. A measure of how much a model drifts apart from equivariance is given by
\begin{equation}
    \label{eq:tilt-sigma}
    \sigma_\textup{rot} = \frac{1}{s} \Biggl(\frac{1}{6} \sum_{k=1}^6 \braket{(c_k - \overline{c})^2}\Bigg)^{1/2},
\end{equation}
where $\overline{c}$ is the mean of the $c_k$ and $\braket{\cdot}$ denotes the area-weighted spatial mean. We calculate $\sigma_\textup{rot}$ for the height $h$ and the vorticity $\zeta$ obtained from the velocity, favoring it over using velocity directly due to its invariance under rotations of the reference frame.  We note that the interpolation, as well as the calculation of $\zeta$ from velocity vectors, results in an interpolation floor, as reported in \autoref{tab:jets-equivariance}.

\begin{wraptable}{r}{0.5\textwidth}
  \centering
  \footnotesize
  \setlength{\tabcolsep}{3pt}
  \caption{Rotation-equivariance error $\sigma_\textup{rot}$ (\autoref{eq:tilt-sigma}) on \galewskytt; 0 for an exactly equivariant model. Floor: $\sigma_\textup{rot}$ of the six realigned ground truths. Best in bold, second best colored green.}
  \label{tab:jets-equivariance}
  \begin{tabular}{@{}l cc cc@{}}
    \toprule
    & \multicolumn{2}{c}{Vorticity $\zeta$} & \multicolumn{2}{c}{Height $h$} \\
    \cmidrule(lr){2-3} \cmidrule(lr){4-5}
    Model & \textit{Next} & \textit{1:20} & \textit{Next} & \textit{1:20} \\
    \midrule
    \textsc{FNO} & 0.1865 & 4.7903 & 0.0380 & 1.5419 \\
    \textsc{SFNO} & 0.1206 & 0.3338 & 0.0059 & 0.0908 \\
    \textsc{GAOT} & 0.4793 & 0.7973 & 0.0252 & 0.1140 \\
    \textsc{LR$^2$Trnsf.} & 0.1917 & 6.4718 & 0.0076 & 0.2132 \\
    \textsc{LS$^2$Trnsf.} & \second{0.0265} & 0.2856 & 0.0052 & 0.0798 \\
    \textsc{Flower2D} & 0.0703 & \second{0.2382} & \textbf{0.0027} & \second{0.0230} \\
    \textsc{\netname} & \textbf{0.0137} & \textbf{0.0743} & \second{0.0028} & \textbf{0.0183} \\
    \midrule
    Floor & \multicolumn{2}{c}{0.0087} & \multicolumn{2}{c}{0.0001} \\
    \bottomrule
  \end{tabular}
\end{wraptable}
We find that on vorticity $\zeta$, \netname has the lowest $\sigma_\textup{rot}$ at step 1 (within $1.6\times$ of the floor), ahead of LS$^2$Trnsf.\ and Flower2D; it is also lowest over the 1:20 rollout ($3.2\times$ below Flower2D). For a visual comparison, see the third row of \autoref{fig:jets-pred}. Height $h$ is overall less discriminating: all models except FNO and GAOT are within a small range for next-step prediction.

Notably, even SFNO does not show equivariance, since it is equivariant by construction only for scalar fields: as the velocity components $u$, $v$ are defined with respect to the local coordinate frame, a rotation mixes them with each other. We expect that SFNO would be equivariant up to discretization error on datasets where fields are independent of the coordinate system (i.e. scalar valued problems such as \mickelintt or \cahnhilliardtt).

\subsection{Rollouts on \mickelintt}
\label{sec:mickelin-analysis}
\begin{figure}[t]
    \centering
    \includegraphics[width=\linewidth]{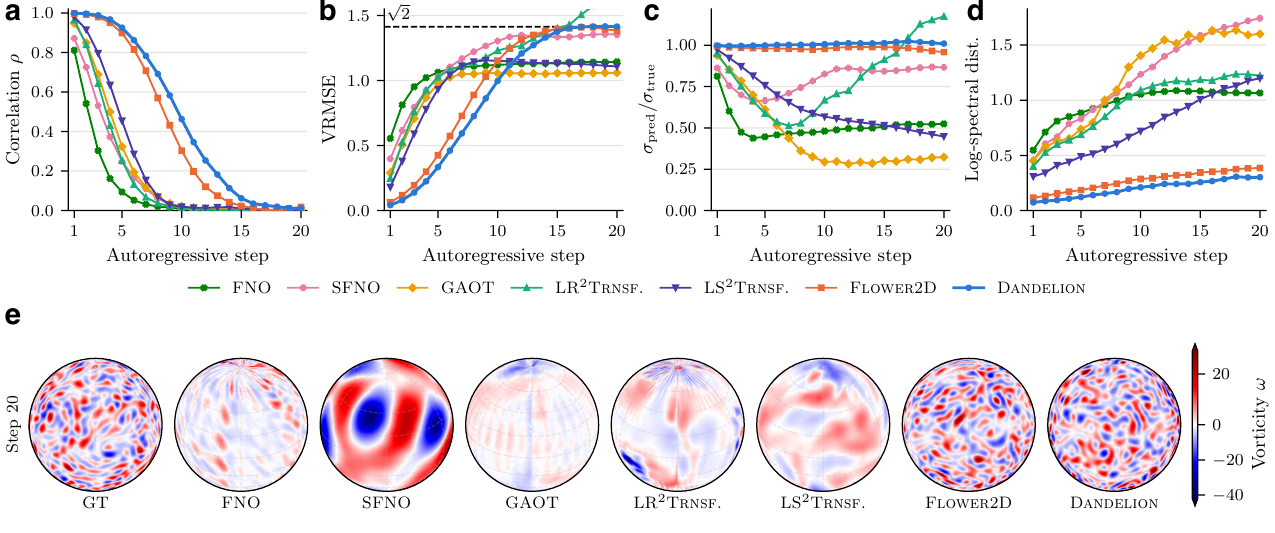}
    \caption{
    Rollout diagnostics on \mickelintt (test-set means). (a) Pearson correlation (area-weighted) of prediction and ground truth. (b) VRMSE; dashed: $\sqrt{2}$, score of a spatially uncorrelated prediction with correct statistics.
    (c) Ratio of predicted to true standard deviation. (d) Log-spectral distance over $\ell=2$--63. (e) Vorticity at step 20 for one test run.
    }
    \label{fig:mickelin-analysis}
\end{figure}

On \mickelintt, \netname has a clear advantage in next-step prediction ($4.7\times$ better than LS$^2$Trnsf.) which shrinks in long-term forecasting (all 1:20 rollout scores lie between $0.89$ and $1.17$). We argue this is due both to the underlying dynamics and the metric used.

Since the system is chaotic, arbitrarily small errors grow exponentially fast, thus making it impossible to track exact trajectories over many steps: for every model other than Flower2D and \netname, the spatial correlation $\rho$ with the reference drops to nearly zero from step $10$ onwards (\autoref{fig:mickelin-analysis}a). Moreover, VRMSE is a pixel-wise metric, rewarding blurring instead of statistical fidelity when correlation is low: to see why, note that, assuming a model predicts the mean correctly $\mu_\textup{pred}=\mu_\textup{true}$, one can decompose $\text{VRMSE}^2 = 1 + a^2 - 2\rho a$, where $\rho$ is again the correlation and $a = \sigma_{\textup{pred}} / \sigma_{\textup{true}}$ measures the relative magnitude of fluctuations about the mean. When $\rho \approx 0$, $\text{VRMSE} = \sqrt{1 + a^2}$ and a prediction damped towards the mean ($a\approx 0)$ results in a lower error than one with the correct variance ($a \approx 1$) but wrong phase. A statistically correct but uncorrelated prediction scores $\sqrt{2}\approx 1.41$, the value both \netname and Flower2D reach at step 20, while architectures producing visibly blurred solutions score lower. This damping is directly visible in the amplitude ratio (\autoref{fig:mickelin-analysis}c): FNO, GAOT and LS$^2$Trnsf.\ drop to $0.3$--$0.5$ of the true standard deviation, while \netname stays at $1.00$. GAOT's second best performance in the \mickelintt rollout column of \autoref{tab:results41} is therefore an artifact of damping combined with a pathology of pixel-wise losses rather than of better predictions. Considering instead the log-spectral distance (\autoref{fig:mickelin-analysis}d), which compares the energy per scale regardless of phase, the models separate clearly with \netname and Flower2D strongly outperforming all other models.

Overall, out of all models only Flower2D and \netname remain statistically plausible, as shown in~\autoref{fig:mickelin-analysis}e: at step 20, they are the only architectures still showing vortex chains comparable to the reference in scale and strength.

\subsection{Discussion of \cahnhilliardtt Results}
\label{sec:ch-analysis}

\begin{table}[t]
  \centering
  \small
  \setlength{\tabcolsep}{3pt}
  \caption{1:20 rollout on \cahnhilliardtt, split by whether the reference field has phase-separated by the last rollout step (28 of 57 test trajectories) or not (29). Mass drift is the mean $|\Delta m/m_0|$ over 20 steps, with its multiple of the solver's own drift in parentheses. Best in bold, second best colored green. FNO is omitted (it diverges).}
  \label{tab:ch-physics}
  \begin{tabular}{@{}l cccc c@{}}
    \toprule
    & \multicolumn{4}{c}{Separated (28)} & Unseparated (29) \\
    \cmidrule(lr){2-5} \cmidrule(l){6-6}
    Model & VRMSE & Mass drift & trajs.\ dissipating & $F$ rises on & VRMSE \\
    \midrule
    GT & -- & \num{4.54e-4} (1\texttimes) & 100\% & 2.7\% & -- \\
    \midrule
    \textsc{SFNO} & 0.935 & \num{2.36e-2} (52\texttimes) & 0\% & 92.1\% & 79.8 \\
    \textsc{GAOT} & 0.711 & \num{3.95e-2} (87\texttimes) & 10.7\% & 65.0\% & \second{28.3} \\
    \textsc{LR$^2$Trnsf.} & 0.767 & \textbf{\num{1.15e-2} (25\texttimes)} & 3.6\% & 86.2\% & 54.9 \\
    \textsc{LS$^2$Trnsf.} & \second{0.654} & \second{\num{1.30e-2} (29\texttimes)} & 64.3\% & 35.4\%  & 51.1 \\
    \textsc{Flower2D} & 0.723 & \num{3.78e-2} (83\texttimes) & \second{85.7\%} & \second{25.9\%} & \textbf{13.5} \\
    \textsc{\netname} & \textbf{0.378} & \num{1.76e-2} (39\texttimes) & \textbf{100\%} & \textbf{5.4\%} & 31.3 \\
    \bottomrule
  \end{tabular}
\end{table}

The \cahnhilliardtt dataset trajectories consist of a fast spinodal transient (phase separation) followed by slow coarsening towards an equilibrium configuration. The separation time depends on the parameters, and out of 57 test trajectories only 28 achieve phase separation within the first 20 steps. We find that models often predict early separation nonetheless: then, in unseparated trajectories, small differences lead to exceptionally high errors, due to the VRMSE being normalized by the reference field's standard deviation. We analyze the two regimes separately: we call a trajectory \emph{separated} if $\max \phi - \min \phi > 0.8$ for the concentration field $\phi$ at the last step, and \emph{unseparated} otherwise. Results are in~\autoref{tab:ch-physics}: for separated trajectories, in addition to VRMSE, we report physics-related metrics to assess mass conservation and free energy monotonicity. More specifically, for mass (which is conserved) we report the mean absolute mass drift over 20 steps, both in absolute terms and as a multiple of the solver's own drift; for the free energy (which is non-increasing) we report the fraction of dissipating trajectories over the full rollout and the fraction of single steps with increasing $F$.

On separated trajectories, \netname achieves the best VRMSE and free energy, measuring how faithfully the geometric structure is reproduced. On mass conservation, it ranks third at $39\times$ the base drift, behind LR$^2$Trsnf.~($25\times$) and LS$^2$Trsnf.~($29\times$); on free energy, it is the only learned model dissipating free energy on every trajectory, and it has the lowest fraction of steps with increasing $F$ ($5.4\%$, with Flower2D ranking second best at $25.9\%$ and the solver baseline being $2.7\%$). We note a discrepancy between pointwise metrics and physical fidelity: LR$^2$Trsnf.~has the lowest mass drift among learned models, yet shows a free energy increase on 27/28 trajectories (the median increase being $231\%$); LS$^2$Trnsf.~has second-best VRMSE and mass drift, but dissipates on only 18/28.

\subsection{Leveraging the third dimension: \netname-2.5D}
\label{sec:2p5d}

\begin{wraptable}{r}{0.56\textwidth}
  \centering
  \footnotesize
  \setlength{\tabcolsep}{3pt}
  \caption{Area-weighted VRMSE of \netname-2.5D on the two 3D datasets vs.\ the channel-stacked models of \autoref{tab:results41} ($\sim$20M parameters each). \netname-2.5D's parameter count scales with the number of vertical levels (18.1M on \heldsuareztt, 10.0M on \oceantt). Best in bold.}
  \begin{tabular}{@{}l cc cc@{}}
    \toprule
    & \multicolumn{2}{c}{\heldsuareztt} & \multicolumn{2}{c}{\oceantt} \\
    \cmidrule(lr){2-3} \cmidrule(lr){4-5}
    Model & \textit{Next} & \textit{1:20} & \textit{Next} & \textit{1:20} \\
    \midrule
    \textsc{Flower2D}        & 0.0290 & 0.2214 & 0.0253 & 0.232 \\
    \textsc{\netname}        & 0.0274 & 0.2067 & 0.0252 & 0.197 \\
    \textsc{\netname-2.5D}   & \textbf{0.0135} & \textbf{0.101} & \textbf{0.0091} & \textbf{0.114} \\
    \bottomrule
  \end{tabular}
  \label{tab:results-2p5d}
\end{wraptable}
Stacking vertical levels along the channel dimension ignores the coupling between neighboring levels, and, while a fully 3D operator could be easily constructed by treating the z-axis as cartesian, it would significantly increase the computational cost of the model. \netname-2.5D is designed as a middle ground. It keeps all horizontal mixing in the 2D tangent warps and SHT resamplings, and carries an explicit level axis that pointwise vertical mixing coarsens and refines together with the horizontal resolution. Such a structure has been used previously by \citet{couairon2024archesweatherefficientaiweather}. Surface and level-wise variables get separate encoders and decoders (\S\ref{app:model-2p5d}). \autoref{tab:results-2p5d} compares it to the channel-stacked models of \autoref{tab:results41}. On \heldsuareztt, the 2.5D variant halves the VRMSE of channel-stacked \netname for both next-step prediction and 1:20 rollout. On \oceantt, it reduces next-step error by $2.8\times$ and rollout error by $1.7\times$. For comparison, \netname improves on Flower2D by at most $1.2\times$ on these two datasets. \netname-2.5D also has fewer parameters than the channel-stacked models: $\sim$18M on \heldsuareztt and $\sim$10M on \oceantt, against $\sim$20M, but it is more expensive to train: on \heldsuareztt, an epoch takes $1.8\times$ as long as for channel-stacked \netname in the single-step phase and $3.3\times$ in the rollout phase.

\section{Conclusion}
\label{sec:conclusion}

We introduced \netname, a spherical adaptation of Flower in which displacements live in the tangent plane and resolution changes pass through a spherical harmonic transform, leaving the warp as the only spatial-mixing operator. A warp-based model is best in every setting of \autoref{tab:results41} and \netname in 13 of 14; the spherical primitives add a clear margin over Flower2D, and the 2.5D variant improves further on both 3D datasets (\S\ref{sec:2p5d}). Our six new spherical PDE datasets (\S\ref{sec:datasets}) complement \texttt{planetswe} and are small enough to train mid-size models from scratch in hours, making it cheap to iterate on spherical architectures before scaling to operational data.

\paragraph{Limitations} Our resolutions cap at $256\times 512$ (approximately $0.7^\circ$), while operational weather models now train at $0.25^\circ$ ($720\times 1440$). Training is short-horizon, with at most two autoregressive steps in the loss, so we do not characterize stability and drift over long rollouts. The benchmark itself is also synthetic. Strong performance on these problems can be seen as a strong indicator, perhaps a necessary condition, but not sufficient for a good weather model.

In the main benchmark we treat the two 3D datasets as 2D fields with the vertical levels stacked along the channel dimension. The 2.5D variant in \S\ref{sec:2p5d} reduces \netname's VRMSE on these datasets by $1.7$--$2.8\times$, so this choice costs accuracy; we have not trained the baselines in a comparable 2.5D setting as this would require a non-trivial amount of engineering. Our main metric is VRMSE, a pointwise error, which can misrank models and ignores physical consistency as seen in \S\ref{sec:mickelin-analysis} and \S\ref{sec:ch-analysis}. Finally, \netname runs on equiangular nodes, on which the discrete SHT is not quadrature-exact at the Nyquist degree, although this does not seem to cause much harm.

\begin{ack}
TM and ID were partially supported by the European Research Council Consolidator Grant 101232534 (PhaseShift). We additionally acknowledge compute resources provided by sciCORE at the University of Basel.
\end{ack}

\printbibliography

\newpage

\appendix
\section{Extended Dataset Descriptions}
\label{app:dataset}

\begin{table}[htbp]
    \centering
        \caption{Summary of the different datasets' main physical features: fields refer to the target variables, physical parameters are the variables modified over different runs. For each combination of given physical parameters, multiple runs are performed with different seeds.}
    \begin{tabular}{l l l}
    \toprule
    Dataset & Fields & Physical parameters \\
    \midrule 
    \galewskytt & \makecell[l]{Zonal~velocity \\ Meridional~velocity \\ Surface height} & \makecell[l]{Maximum jet velocity \\ Jet mid latitude \\ Perturbation amplitude \\ Global average depth} \\
    \midrule
    \mickelintt & Vorticity & \makecell[l]{Normalized radius \\ Normalized bandwidth} \\
    \midrule 
    \cahnhilliardtt & Order parameter & \makecell[l]{Interface width \\ Mean composition \\ IC noise variance \\ Sphere radius} \\
    \midrule 
    \shockcapstt & \makecell[l]{Fluid depth \\ Zon.~depth-integrated momentum \\ Mer.~depth-integrated momentum} & \makecell[l]{Number of caps \\ Velocity scaling} \\
    \midrule 
    \heldsuareztt & \makecell[l]{Zonal velocity \\ Meridional velocity \\ Air temperature \\ Surface pressure} & \makecell[l]{Angular velocity \\ Mer. temperature delta \\ Vertical temperature delta} \\ 
    \midrule
    \oceantt & \makecell[l]{Potential temperature \\ Salinity \\ Face-$x$ velocity \\ Face-$y$ velocity \\ Sea-surface height} & \makecell[l]{GM/Redi background diffusivity \\ Horizontal viscosity \\ Vertical diffusivity \\ Surface temperature restoring \\ Surface salinity restoring} \\ 
    \bottomrule 
    \end{tabular}

    \label{tab:datasetsrecap}
\end{table}

In this Appendix, we provide detailed specifications for the spherical PDE datasets introduced in this work. A summary of the main physical quantities can be found in~\autoref{tab:datasetsrecap}.

\subsection{\texttt{planetswe}}

\paragraph{Underlying physics} The \texttt{planetswe} dataset~\cite{mccabe2023towards} is the only dataset in our suite that we did not generate ourselves; it ships as part of The Well collection~\cite{ohana2025welllargescalecollectiondiverse}. It solves a forced, hyper-viscous shallow water system on a rotating unit sphere,
\begin{align}
    \partial_t \vec{u} &= -\vec{u}\cdot\nabla\vec{u} - g\nabla h - \nu\nabla^4\vec{u} - 2\vec{\Omega}\times\vec{u}, \\
    \partial_t h       &= -H\nabla\cdot\vec{u} - \nabla\cdot(h\vec{u}) - \nu\nabla^4 h + F,
\end{align}
where $h$ is the deviation of the pressure-surface height from a mean reference height $H$, $\vec{u}$ the 2-D velocity tangent to the sphere, $\vec{\Omega}$ the Coriolis vector, and $F$ a spatially localised heat source that drives the system on a non-trivial seasonal cycle. The hyper-viscous $\nu\nabla^4$ term is normalised against spherical-harmonic mode $224$. The forcing $F$ is concentrated at the day/night terminator and its antipodal point: its longitude centre varies with the time of day (daily cycle) and its latitude centre is modulated sinusoidally over the year (annual cycle, with a maximum solar declination of $\sim 0.4$\,rad and an angular width $\sigma = \pi/2$). The combination of explicit Coriolis, a slow seasonal forcing, and high-order dissipation produces a non-autonomous flow with strong day-of-year structure, which is the property the dataset is designed to test.

\paragraph{Simulation details} Trajectories are integrated with the Dedalus spectral framework\,\cite{dedalus:framework}, adapted from its shallow-water-on-sphere example. Spatial discretisation is a $256 \times 512$ equiangular grid in polar coordinates with $\phi \in [0, 2\pi]$ and $\theta \in [0, \pi]$. The time step is CFL-adaptive with safety factor $0.4$, and snapshots are written every hour of simulated time. Each trajectory is initialised from the $500$-hPa level of an ERA5 frame, filtered for numerical stability, and burned in for half a simulated year before output begins. Each retained trajectory then spans three model years, i.e.~$3024$ snapshots per trajectory ($24$ snapshots per day, $1008$ snapshots per year). One simulation takes roughly $45$ minutes on $64$ Icelake cores.

\paragraph{Fields} Available fields are the surface-height deviation $h$ (scalar) and the two components of the tangent velocity $\vec{u}$, in the conventions used by The Well release.

\paragraph{Ensemble} The published dataset comprises $40$ trajectories, distinguished only by their ERA5-derived initial condition; there is no parameter sweep over physical constants. The total on-disk ensemble size is $185.8$\,GB. We use the train/val/test split provided by The Well unmodified.

\paragraph{References} \cite{mccabe2023towards,ohana2025welllargescalecollectiondiverse,dedalus:framework}.

\subsection{\galewskytt}
\label{app:datasets-galewsky}
\begin{figure}[t]
    \centering
    \includegraphics[width=\linewidth]{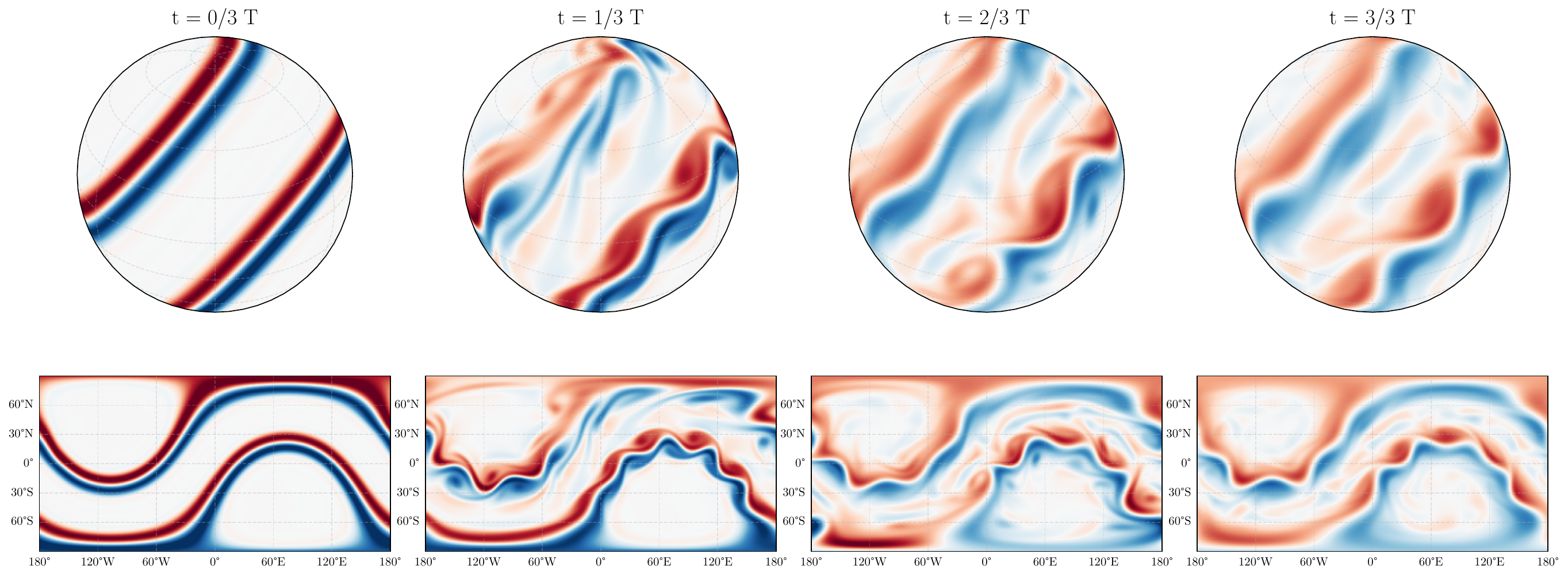}
    \caption{Evolution of the vorticity along a \galewskytt trajectory. Top row: states depicted on the sphere. Bottom row: states depicted on a latlon grid. Color scale limits (vmin/vmax) are shared. } 
    \label{fig:galewsky-4-evolution}
\end{figure}

\begin{table}
    \centering
    \caption{Grid of varying physical and stochastic parameters across the spherical PDE datasets.}
    \begin{tabular}{llllc}
    \toprule
    Dataset & Parameter & Units & Values & Count \\
    \midrule
    \multirow{5}{*}{\galewskytt}
        & $u_\textup{max}$    & $\si{\meter\per\second}$  & $60, 70, 80, 90, 100$       & 5 \\
        & $\phi_\textup{mid}$ & \textdegree N             & $30, 40, 50, 60$            & 4 \\
        & $h_\textup{pert}$   & $\si{\meter}$             & $60, 240$                   & 2 \\
        & $\overline{h}$      & $\si{\meter}$             & $8000, 10000, 12000, 14000$ & 4 \\
        & Seed                & --                        & $0, 1, 2, 3, 4, 5$          & 6 \\
    \midrule
    \multirow{3}{*}{\mickelintt}
        & $R/\Lambda$      & -- & $2, 4, 7, 10$                 & 4 \\
        & $\kappa\Lambda$  & -- & $0.4, 0.7, 1.0, 1.4, 1.8$     & 5 \\
        & Seed             & -- & $0, 1, \ldots, 39$            & 40 \\
    \midrule
    \multirow{5}{*}{\cahnhilliardtt}
        & $\varepsilon$              & -- & $0.5, 1.0, 1.5, 2.0$       & 4 \\
        & $\overline{\phi}_0$     & -- & $0.35, 0.50, 0.65$         & 3 \\
        & $\sigma_0^2$               & -- & $0.001, 0.005, 0.01, 0.05$ & 4 \\
        & $R$                        & -- & $5.0, 7.5, 10.0$           & 3 \\
        & Seed                       & -- & $0, 1, 2, 3$               & 4 \\
    \midrule
    \multirow{3}{*}{\shockcapstt}
        & $K$        & -- & $1, 2, 4, 8, 16$              & 5 \\
        & $\delta$   & -- & $0.0, 0.25, 0.5, 0.75, 1.0$   & 5 \\
        & Seed       & -- & $0, 1, \ldots, 19$            & 20 \\
    \midrule 
    \multirow{4}{*}{\heldsuareztt}
        & $\Omega/\Omega_\textup{earth}$ & -- & $0.5, 1.0, 2.0$ & 3 \\
        & $\Delta T_y$ & $\si{\kelvin}$ & $40, 60, 80$ & 3 \\ 
        & $\Delta \theta_z$ & $\si{\kelvin}$ & $5, 10, 20$ & 3 \\ 
        & Seed & -- & $0, 1, 2, 3, 4, 5$ & 6 \\ 
    \midrule 
    \multirow{5}{*}{\oceantt}
        & $\kappa_\textup{GM}$ & $\si{\meter\squared\per\second}$ & $250, 1000, 2500$ & 3 \\ 
        & $A_h$ & $\si{\meter\squared\per\second} $ & $\num{1.5e5}, \num{3e5}, \num{5e5}$ & 3 \\ 
        & $K_v$ & $\si{\meter\squared\per\second}$ & $\num{1e-5}, \num{3e-5}, \num{1e-4}$ & 3 \\ 
        & $\tau_T$ & days & $30, 60, 120$ & 3 \\ 
        & $\tau_S$ & days & $90, 180, 360$ & 3 \\ 
    \bottomrule
    \end{tabular}
    
    \label{tab:datasetparams}
\end{table}

\paragraph{Underlying physics} The Galewsky problem~\cite{galewsky:paper} is a standard test for numerical solvers, describing the shallow water equations on a sphere of radius $R$, rotating with angular velocity $\Omega$:
    \begin{align}
        \frac{\mathrm{D}\vec{V}}{\mathrm{D}t} &= - f \hat{\vec{k}}\times \vec{V} - g\nabla h + \nu \nabla^2 \vec{V}, \\
        \frac{\mathrm{D}h}{\mathrm{D}t} &= -h \nabla \cdot \vec{V} + \nu \nabla^2 h, 
    \end{align}
where $\mathrm{D}/\mathrm{D}t$ denotes the material derivative, $\vec{V}=u\hat{\vec{i}} + v\hat{\vec{j}}$ is the velocity tangent to the surface, $\hat{\vec{i}}$, $\hat{\vec{j}}$ and $\hat{\vec{k}}$ are the unit vector in the eastward, northward, and normal direction respectively, $f \equiv 2 \Omega \sin\phi$ is the Coriolis frequency ($\phi$ being the latitude), $g$ is the gravitational acceleration, $\nu$ is the diffusion coefficient, and $h$ is the depth of the fluid. Parameters are set to the standard values $R = \SI{6.371e6}{\meter}$, $\Omega = \SI{7.292e-5}{\per\second}$, $g = \SI{9.80616}{\meter\per\second\squared}$, and $\nu = \SI{1e-5}{\meter\squared\per\second}$.

In the standard Galewsky setting, the initial conditions are analytically specified to ensure reproducibility, and correspond to a mid-latitude zonal flow and a balanced height field, with a localized perturbation to develop barotropic instability. In particular, the flow velocity is 
    \begin{equation}
        u(\phi) = \begin{dcases}
            0 &\phi \le \phi_0 \\
            \frac{u_\textup{max}}{e_\textup{n}}\exp\biggl[\frac{1}{(\phi - \phi_0)(\phi - \phi_1)}\biggr] &\phi_0 < \phi < \phi_1 \\
            0 &\phi > \phi_1
        \end{dcases}
    \end{equation}
where $u_\textup{max}$ is the maximum velocity, $\phi_0$ (resp.~$\phi_1$) the southern (northern) boundary latitude of the flow in radians, and $e_\textup{n}$ a dimensionless rescaling parameter to ensure the flow velocity reaches $u_\textup{max}$ at the midpoint. The absence of local and advective acceleration simplifies the balance equation to 
    \begin{equation}
        fu + \frac{u^2(\phi)\tan(\phi)}{R} = - \frac{g}{R}\frac{\partial h}{\partial \phi},
    \end{equation}
from which the height can be obtained via numerical integration: 
    \begin{equation}
        g h(\phi) = gh_0 - \int^\phi Ru(\phi') \biggl[f + \frac{u(\phi')\tan(\phi')}{R}\biggr]\,\mathrm{d}\phi', 
    \end{equation}
where $h_0$ is chosen to fix the value of the global average depth $\overline{h}$. Finally, the height perturbation has the form 
    \begin{equation}
        \Delta h(\lambda, \phi) = h_\textup{pert} \cos(\phi)\, \mathrm{e}^{-(\lambda/\alpha)^2}\, \mathrm{e}^{-\bigl[(\phi_2 - \phi)/\beta\bigr]^2}, \quad -\pi < \lambda < \pi,
    \end{equation}
where $\lambda$ is the longitude, $\phi_2 = \pi/4$, $\alpha = 1/3$, and $\beta = 1/15$. 

We extend this setting with two additions: 
\begin{enumerate*}[label=(\arabic*)]
    \item we include a mirrored zonal jet in the southern hemisphere \emph{without} the height perturbation, resulting in Rossby waves from the first jet triggering a barotropic instability in the second one; 
    \item as a post-processing step, we randomly sample a rotation matrix $R\in\mathrm{SO}(3)$ and apply a spatial pullback $R^*$ to transform the coordinates and the fields, obtaining a rotated version of the simulation. 
\end{enumerate*}
The resulting dataset can be challenging for autoregressive models, with the second triggered instability acting as a diagnostic test for excessive numerical dissipation, and the added rotation forcing the model to learn the proper physics from the dynamical evolution rather than just memorizing a simple mapping with the geometrical grid.

\paragraph{Simulation details} The simulation is carried out through MPI-parallel spectral methods using the Dedalus v3 framework, using a second order 2-stage DIRK+ERK solver scheme (RK222). The integration was performed on a Gauss-Legendre (in colatitude) and equispaced (in longitude) grid with $N_\theta = 256$, $N_\phi = 512$, and a dealias factor of $3/2$; snapshots are then resampled to a regular lat-lon grid with cubic-spline interpolation. To ensure numerical stability, an adaptive CFL time step, initialized at $(\Delta t)_\textup{in} = \SI{120}{\second}$ and capped at $(\Delta t)_\textup{max} = \SI{600}{\second}$, was employed. Each simulation spans $32$ days of simulated time, with snapshots being taken every $4$ hours of simulated time; the first $4$ days of linear spin-up were discarded, resulting in a total of $168$ snapshots per trajectory. The simulation is executed in \texttt{float64} precision, then downcast to \texttt{float32} for storage. Generating the full dataset required approximately $\num{10000}$ to $\num{13000}$ core-hours on the sciCORE cluster (AMD Epyc/Intel Xeon nodes), with individual runs taking roughly 40 to 50 minutes using 16 MPI ranks. 

\paragraph{Fields} Available fields are the zonal velocity $u$, the meridional velocity $v$, and surface-height perturbation $h$ (cf.~\autoref{tab:datasetsrecap}). 

\paragraph{Initial conditions}
The IC for each run is the analytic Galewsky construction extended to two hemispheres, parameterised by $(u_{\max}, \varphi_{\mathrm{mid}}, h_{\mathrm{pert}}, \bar{h})$ from the sweep grid: a bi-hemispheric zonal-jet velocity field summing the canonical northern jet at $+\varphi_{\mathrm{mid}}$ with peak amplitude $u_{\max}$ and a mirrored southern jet at $-\varphi_{\mathrm{mid}}$ (Eq.~9); the geostrophically- and cyclostrophically-balanced surface-height profile $h(\varphi)$ obtained by 1-D numerical integration of the zonal balance equation) followed by cos-weighted area-mean subtraction enforcing $\langle h \rangle = 0$; and a localised bi-Gaussian height perturbation of amplitude $h_{\mathrm{pert}}$ applied only on the northern jet to break hemispheric symmetry. The IC is constructed and the simulation runs in the canonical (polar-aligned) frame, then a per-trajectory rotation $R(\hat{\mathbf{e}}, \alpha) \in \mathrm{SO}(3)$, with axis $\hat{\mathbf{e}}$ uniform on $S^2$ and angle $\alpha$ uniform on $[0, 2\pi)$, is applied at postprocess time to the resampled $(\mathrm{lat}, \mathrm{lon})$ snapshots. The rotation pair is keyed on the run identifier rather than on the \texttt{seed} value, so all $960$ trajectories receive distinct tilts and the \texttt{seed} axis serves purely as a rotation-multiplicity index over each $(u_{\max}, \varphi_{\mathrm{mid}}, h_{\mathrm{pert}}, \bar{h})$ combination.

\paragraph{Parameter grid} Varying physical parameters are the maximum jet velocity $u_\textup{max}$, the jet mid latitude $\phi_\textup{mid} = (\phi_1 - \phi_0)/2$, the perturbation amplitude $h_\textup{pert}$, and the global average depth $\overline{h}$. For each run, six random rotation matrices were applied, to augment the dataset with spatially rotated version of each simulation. Detailed values can be found in~\autoref{tab:datasetparams}.

\paragraph{References} \cite{galewsky:paper,dedalus:framework}.

\subsection{\mickelintt}
\begin{figure}[t]
    \centering
    \includegraphics[width=\linewidth]{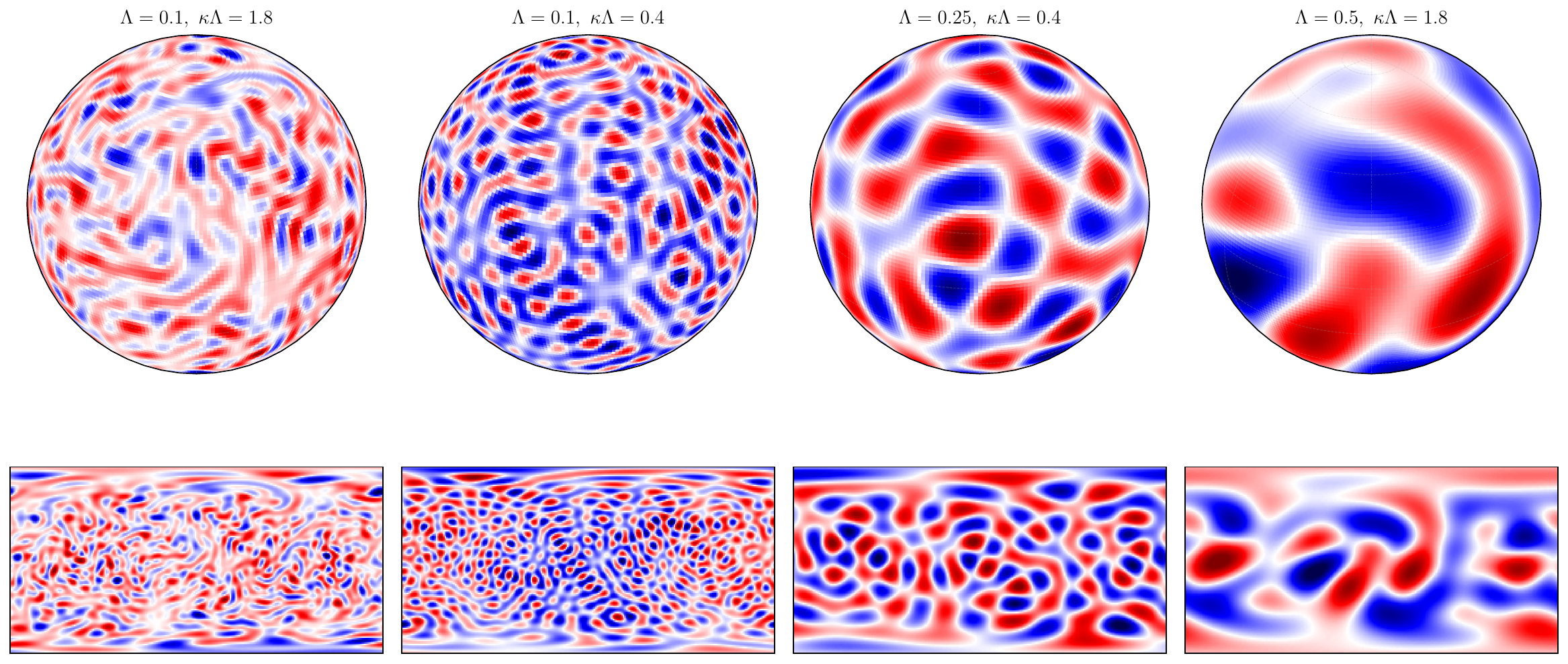}
    \caption{Four different final states of trajectories from \mickelintt, representing the corners of the parameter space. Top row: states depicted on the sphere. Bottom row: states depicted on a latlon grid. Color scale limits (vmin/vmax) differ by plot, as different phases reach different limits.} 
    \label{fig:mickelin-4-samples}
\end{figure}

\paragraph{Underlying physics} We consider the covariant extension of a generalized Navier-Stokes (GNS) for an incompressible active fluid on arbitrarily curved surfaces.~\cite{mickelin:paper} The equations governing such a system are 
    \begin{align}
        \label{eqn:incompr}
        \nabla_a v^a &= 0, \\ 
        \label{eqn:cauchy}
        \partial_t v^a + v^b \nabla_b v^a &= \nabla^a \sigma + \nabla_b T^{ab},
    \end{align}
where $v^a$ are the velocity field components, $\nabla_b v^a$ is the covariant derivative, and $T^{ab}$ is the stress tensor; Eqs.~\labelcref{eqn:incompr,eqn:cauchy} represent respectively incompressibility and Cauchy momentum conservation. The active stress is modeled by the covariant linear constitutive relation
    \begin{align}
            \label{eqn:stresstensor}
            T^{ab} &= f(\nabla^2)(\nabla^a v^b + \nabla^b v^a), \\
            \label{eqn:polyansatz}
            f(\nabla^2) &= \Gamma_0 - \Gamma_2 \nabla^2 + \Gamma_4 \nabla^2\nabla^2,
    \end{align}
where $\nabla^2 = \nabla^a \nabla_a$ is the tensor Laplacian. With the polynomial ansatz for the driving term of Eq.~\eqref{eqn:polyansatz}, for $\Gamma_2 < 0$ the system shows a characteristic bandwidth $\kappa$ of linearly unstable modes characterized by vortices of size $\Lambda$ and growth time $\tau$. Note that, for a sphere of radius $R$, fixing the parameters scale of the problem $(\tau, \Lambda, \kappa)$ uniquely determines the parameters $(\Gamma_0, \Gamma_2, \Gamma_4)$; additional details are in the Supplemental Material of~\cite{mickelin:paper}. 
Nonstationary solutions of Eqs.~\labelcref{eqn:incompr,eqn:cauchy,eqn:stresstensor,eqn:polyansatz} fall into three regimes: the $B$ phase, for $\kappa R \lesssim 1$, is characterized by cycles of quasistationary flow patterns and sudden energy bursts; the $A$ phase, for $R^{-1} < \kappa < \Lambda^{-1}$, shows an anomalous turbulence, featuring self-organizing vortex chains of anti-ferromagnetic order; the $T$ phase, for $\kappa \Lambda > 1$, is an ordinary Kolomogorov turbulence regime. 

The dataset we provide contains the $A$ and $T$ phases only, for statistical and computational reasons: such phases are characterized by continuous spatial evolution which provide dense, high-frequency dynamical information per temporal snapshot, allowing models to efficiently converge on robust representations of turbulent advection without severe ergodic sampling bottlenecks. In contrast, the $B$ phase is highly intermittent, governed by prolonged quasistationary periods abruptly punctuated by sudden energy bursts. This extreme temporal non-stationarity yields heavy-tailed distributions where dynamically informative events are sparse.

\paragraph{Simulation details} The simulation is carried out through MPI-parallel spectral methods using the Dedalus v3 framework, using a second order 2-stage DIRK+ERK solver scheme (RK222). The integration was performed on a Gauss-Legendre (in colatitude) and equispaced (in longitude) grid with $N_\theta = 128$, $N_\phi = 256$, and a dealias factor of $3/2$; snapshots are then resampled to a regular lat-lon grid with cubic-spline interpolation. To ensure numerical stability, an adaptive CFL time step, initialized at $(\Delta t)_\textup{in} = \num{5e-3}$ and capped at $(\Delta t)_\textup{max} = \num{5e-2}$, was employed. Each simulation spans $65\tau$, where $\tau$ is the inverse peak driving rate, with snapshots being taken every $\num{0.2}\tau$. The first $20\tau$, covering the growth of the seed perturbation to nonlinear saturation, are discarded as spin-up, leaving $225$ snapshots per trajectory ($20\tau \le t \le 64.8\tau$). The simulation is executed in \texttt{float64} precision, then downcast to \texttt{float32} for storage. Generating the full dataset required approximately $\num{1000}$ to $\num{3000}$ core-hours on the sciCORE cluster (AMD Epyc/Intel Xeon nodes), with individual runs taking roughly 3 to 20 minutes using 16 MPI ranks. 
 
\paragraph{Fields} The only available field is vorticity $\boldsymbol{\omega}$ (cf.~\autoref{tab:datasetsrecap}). Note that velocity can be obtained by vorticity by solving the spherical Poisson equation for the stream function ($\Delta\psi = -\omega$), and then computing the skew-gradient of the result. Models that need velocity should add a learned stream-function head or perform the elliptic solve at load time.

\paragraph{Initial conditions}
Each run is seeded with a small-amplitude band-limited random vorticity field $\omega_0(\theta, \varphi) = \varepsilon \sum_{\ell = 2}^{\ell_{\mathrm{init}}} \sum_{m = -\ell}^{\ell} a_{\ell m}\, Y_{\ell m}(\theta, \varphi)$, with i.i.d.~Gaussian coefficients $a_{\ell m} \sim \mathcal{N}(0, 1)$ drawn from \texttt{np.random.Generator(np.random.PCG64(seed))} and amplitude $\varepsilon = 10^{-3}$, well inside the linear regime so that the saturation amplitude is set by the nonlinear balance of drive and dissipation rather than by the seed scale. The constant ($\ell = 0$) and rigid-rotation ($\ell = 1$) modes are excluded as dynamically irrelevant gauges. The band-limit $\ell_{\mathrm{init}} = \lceil R \cdot (\pi/\Lambda + \kappa / 2) \rceil + 4$ is set per run a few harmonic degrees above the upper edge of the unstable band, so the IC seeds directly inside the actively unstable modes and saturation is reached on an $O(\tau)$ timescale rather than after many thousands of solver steps spent waiting for round-off to bootstrap the instability.

\paragraph{Parameter grid} Varying physical parameters are the rescaled radius $R/\Lambda$ and the active bandwidth $\kappa\Lambda$. Physics constant $R=1$ and $\tau=1$ are fixed across the ensemble; the per-run derived parameters are $\Lambda$, $\kappa$, and $(\Gamma_0, \Gamma_2, \Gamma_4)$, from the closed-form coefficient map. Each run is repeated across $40$ random seeds. Detailed values can be found in~\autoref{tab:datasetparams}. 

\paragraph{References} \cite{mickelin:paper,dedalus:framework}

\subsection{\cahnhilliardtt}
\label{app:datasets-cahn-hilliard}
\begin{figure}[t]
    \centering
    \includegraphics[width=\linewidth]{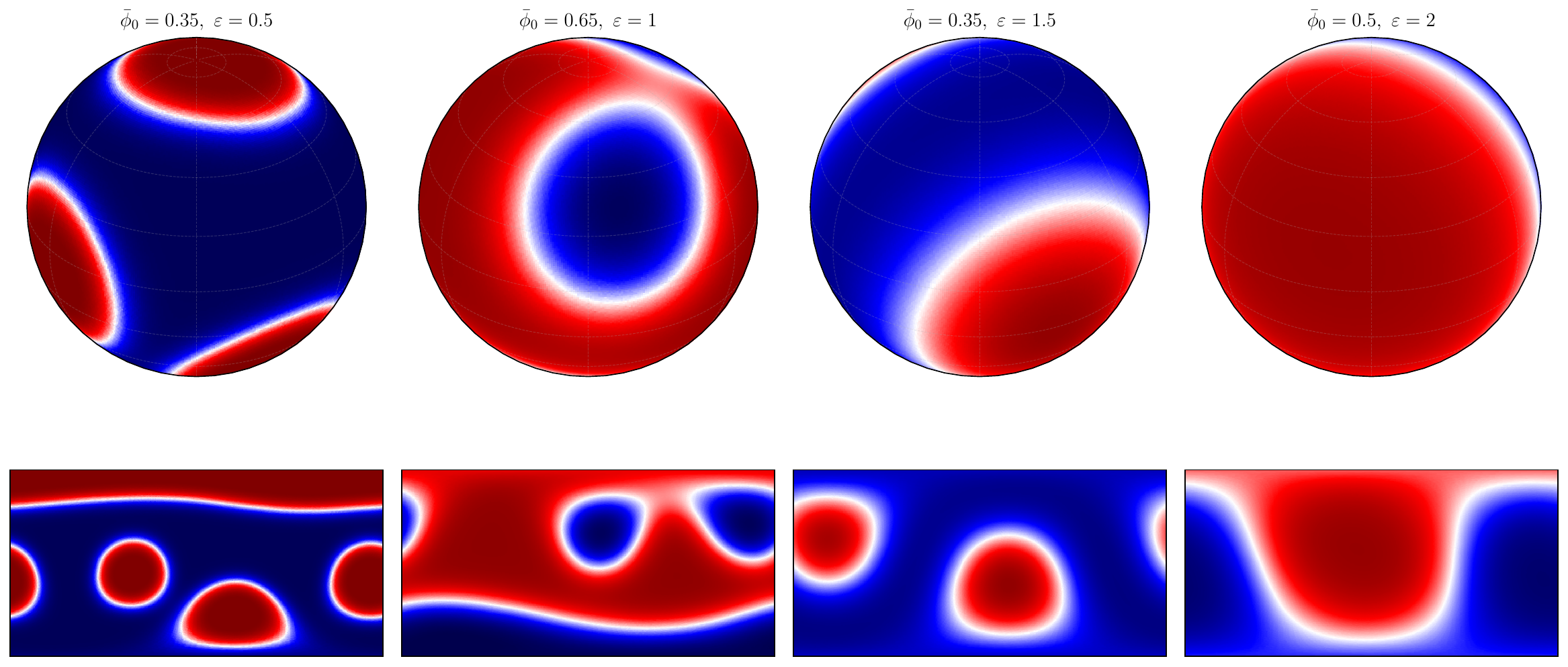}
    \caption{Four different final states of trajectories from \cahnhilliardtt, representing the corners of the parameter space. Top row: states depicted on the sphere. Bottom row: states depicted on a latlon grid. Color scale limits (vmin/vmax) are shared. }
    \label{fig:cahn-hilliad-4-samples}
\end{figure}

\paragraph{Underlying physics} The Cahn-Hilliard equation~\cite{cahnhilliard:paper,cahn:spinoidal} models phase separation by spinodal decomposition, with an order parameter $\phi \in [0, 1]$ interpolating between two coexisting phases at $\phi \approx 0$ and $\phi \approx 1$, separated by diffuse interfaces of width set by $\varepsilon$. We solve the  equation on a sphere of radius $R$:
\begin{equation}
    \frac{\partial \phi}{\partial t} = D \nabla^2 \biggl[a^2 \frac{\partial f}{\partial \phi} - \varepsilon^2 \nabla^2 \phi\biggr] = \nabla\cdot Da^2 \bigl[1 - 6\phi(1 - \phi)\bigr]\nabla\phi - \nabla\cdot D\nabla\varepsilon^2 \nabla^2\phi,
\end{equation} 
where $D$ is the diffusion coefficient and $f = \tfrac12\phi^2(1 - \phi)^2$ a double-well free energy function. The interplay between $f$ (which penalizes intermediate values of $\phi$) and the gradient energy contribution $\tfrac{\varepsilon^2}{2}|\nabla\phi|^2$ (which penalizes steep spatial variations of $\phi$) results in the separation of homogeneous regions in which $\phi = 0$ or $\phi = 1$, separate by narrow yet continuous interfaces, see~\autoref{fig:cahn-hilliad-4-samples}. 

\paragraph{Simulation details}
The simulation is carried out through the NIST FiPy finite-volume framework~\cite{FiPy:2009}, using the \texttt{examples/cahnHilliard/sphere.py} recipe with a first-order implicit time stepper performing one linear solve per step on the coupled diffusion form \texttt{TransientTerm = DiffusionTerm} $-$ \texttt{DiffusionTerm}. The integration was performed on an unstructured \texttt{Gmsh2DIn3DSpace} mesh of the sphere surface (\texttt{cell\_size = 0.3}, six surface patches stitched together for well-conditioned meshing) extruded radially by a factor $1.1$ to a one-cell-thick shell, giving the face-based fluxes a well-defined orientation; cell count grows quadratically with the sphere radius $R$. Snapshots are then resampled to a regular $256 \times 512$ lat-lon grid by inverse-distance weighting on the $k = 4$ nearest cell centres on the unit sphere. To resolve the multi-scale dynamics, an exponential timestep schedule $\Delta t = \min(\Delta t_{\max}, e^{d_{\exp}})$ was employed, with $d_{\exp}$ initialised at $-5$ and incremented by $0.01$ per step, capped at $\Delta t_{\max} = 100$: tiny steps resolve the violent spinodal-decomposition transient near $t = 0$, then $\Delta t$ saturates at the ceiling once coarsening becomes the dominant dynamic. Each simulation spans $630$ solver-time units with snapshots being taken every $\approx10$ solver-time units ($64$ snapshots per trajectory); no spin-up is discarded, since the $t = 0$ Gaussian-noise IC and the spinodal-decomposition transient are part of the dynamics the dataset is designed to capture. The simulation is executed in \texttt{float64} precision, then downcast to \texttt{float32} for storage. Generating the full dataset required approximately $50$ to $250$ core-hours on the sciCORE cluster (AMD Epyc/Intel Xeon nodes), with individual runs taking roughly $5$ to $30$ minutes single-process (FiPy is not MPI-parallelised on this mesh).

\paragraph{Fields} Available field is the order parameter $\phi$ (cf.~\autoref{tab:datasetsrecap}). Small transient overshoots of $\phi$ outside $[0, 1]$ are expected during interface formation and relax as interfaces sharpen.

\paragraph{Initial conditions} Each run is seeded with a per-cell Gaussian field $\phi_0(c) \sim \mathcal{N}(\overline{\phi}_0, \sigma_0^2)$, $c \in \text{mesh cells}$, drawn from \texttt{np.random.default\_rng(seed)} with one i.i.d.~sample per mesh cell. The IC is therefore unstructured and contains no spectral truncation. The mean $\overline{\phi}_0$ and variance $\sigma_0^2$ are both swept axes.

\paragraph{Parameter grid} Varying parameters are the interface width $\varepsilon$, the mean composition $\overline{\phi}_0$, the IC noise variance $\sigma_0^2$, and the sphere radius $R$. Detailed values are in \autoref{tab:datasetparams}. The $\overline{\phi}_0$ axis spans symmetric ($0.50$) and asymmetric ($0.35, 0.65$) compositions, all inside the spinodally-unstable window. The $\varepsilon$ and $R$ axes together control the dimensionless pattern count ($O(R/\varepsilon)$) and span an order-of-magnitude range. The $\sigma_0^2$ axis ranges from a barely-perturbed near-uniform IC ($\sigma_0^2 = 0.001$) to a strongly-perturbed IC ($\sigma_0^2 = 0.05$) that already contains domain-scale structure at $t = 0$.

\paragraph{References} \cite{cahnhilliard:paper, cahn:spinoidal, FiPy:2009}

\subsection{\shockcapstt}
\begin{figure}[t]
    \centering
    \includegraphics[width=\linewidth]{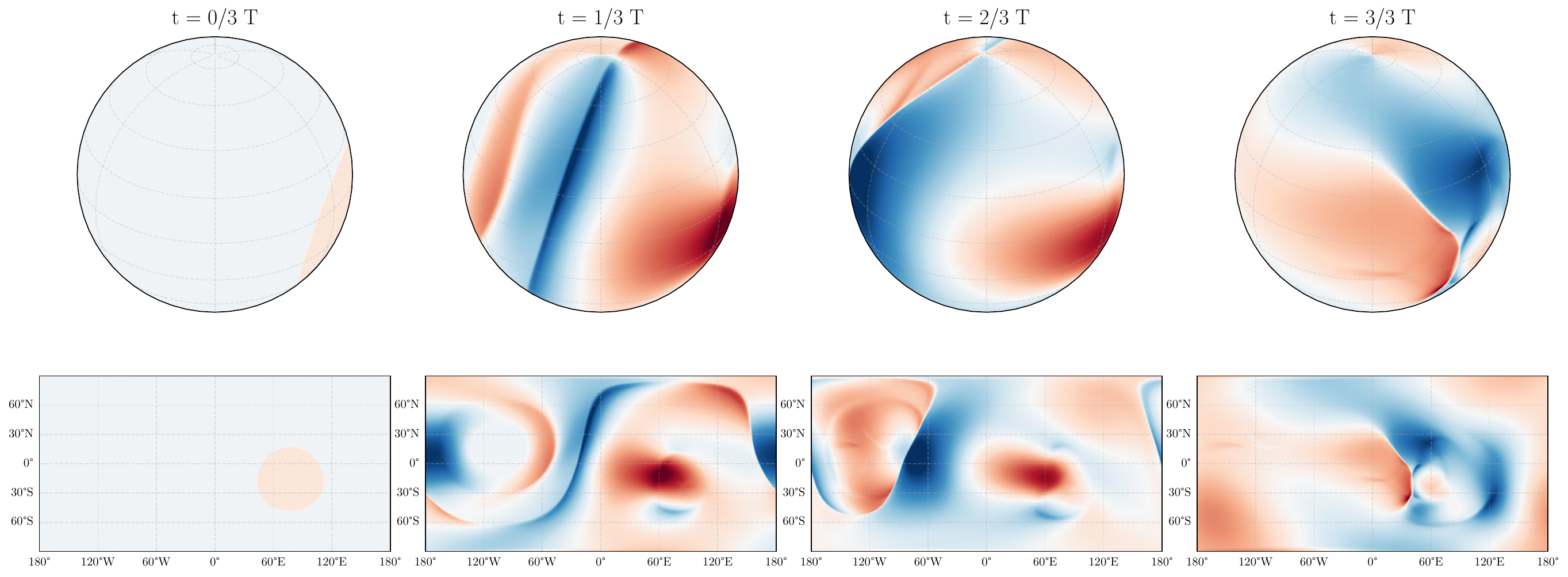}
    \caption{Evolution of the Momentum U-variable of a \shockcapstt trajectory. Top row: states depicted on the sphere. Bottom row: states depicted on a latlon grid. Color scale limits (vmin/vmax) are shared. Note that the radial lines emanating from the poles are not a visual artifact or a solver bug. Because the local eastward basis vector abruptly reverses direction across the pole, the projection of the continuous physical momentum vector onto this basis results in a sharp sign-flip.}
    \label{fig:shock-caps-4-evolution}
\end{figure}

\paragraph{Underlying physics} The system is the rotation-free shallow water equations on the unit sphere $S^2$, written in conservation form for fluid depth $h$ and depth-integrated momentum $h\vec{u}$:
\begin{align}
    \frac{\partial h}{\partial t} + \nabla_{S^2} \cdot (h\vec{u}) &= 0, \\
    \frac{\partial (h\vec{u})}{\partial t}  + \nabla_{S^2} \cdot \biggl(h\vec{u}\otimes\vec{u} + \frac{1}{2} g h^2 \mathbb{I}\biggr) &= 0,
\end{align}
with non-dimensional gravity $g = 9.80616$ and unit sphere radius. The gravity-wave celerity $c = \sqrt{gh}$ closes the characteristic structure; with $h \in [0.5, 2.0]$, $c \in [2.21, 4.43]$. This is fast enough that shock fronts traverse a large fraction of the sphere within the simulated horizon $t_\textup{max} = 1.5$. No Coriolis term is included; the dataset is intentionally rotation-free so the shock physics dominates. Unlike spectral solvers, which require hyperviscosity to stabilise and smear shocks, the dissipation needed to enforce the entropy condition is provided organically by the finite-volume Riemann solver and TVD limiters.

\paragraph{Simulation details} The simulation uses the Clawpack/PyClaw high-resolution finite-volume framework~\cite{clawpack, mandli2016clawpack}, with the \texttt{riemann.shallow\_sphere\_2D} approximate Riemann solver paired with the \texttt{classic2\_sw\_sphere} Fortran step module. Spatial discretisation is the Calhoun--Helzel single-patch mapped sphere~\cite{calhoun2008logically}, a logically rectangular grid covering $S^2$ with $(N_x, N_y) = (512, 256)$ cells over the computational rectangle $[-3, 1] \times [-1, 1]$; this single-patch wrap has no coordinate singularity at the poles, at the cost of only $\sim 4$ source cells (one per logical quadrant) meeting at each pole. Time integration is explicit with the MC limiter, 2-D transverse-wave correction, and a Fortran source-split step (\texttt{sw\_sphere\_problem.src2}) that projects momentum back onto the local tangent plane every macro-step. The CFL is held at $0.45$ with a hard ceiling of $0.9$. Snapshots are remapped to a regular $256 \times 512$ lat-lon grid via a first-order conservative spherical-polygon area-overlap operator: each output cell is the spherical-area-weighted average of the source cells overlapping it, with overlap areas evaluated on the sphere via Girard's theorem with great-circle edges. The remap is mass-conservative, monotone, and shock-preserving. Cartesian momentum components are remapped first, and the projection onto local east/north happens once per target pixel, so basis-rotation artifacts near the poles do not contaminate the area average. Each simulation spans $1.5$ non-dimensional time units, with $101$ snapshots per trajectory at $\Delta t = 0.015$. Per-run wall is roughly $10$ to $30$ minutes (OMP-parallel, $16$ threads), with $K = 16$ runs the budget driver because extra shock fronts tighten the CFL early; the full $500$-run ensemble took $\approx 100$ to $250$ core-hours.

\paragraph{Fields} Available fields are fluid depth $h$ and the zonal and meridional depth-integrated momenta $hu$, $hv$ (cf.~\autoref{tab:datasetsrecap}). The Fortran solver carries 3-D Cartesian momentum internally; the four Cartesian components $(h, hu_x, hu_y, hu_z)$ are remapped to the lat-lon grid first, and only then projected to local east/north using the target grid's coordinates -- doing the projection before remapping would silently cancel real momentum near the poles, where the east/north basis rotates rapidly.

\paragraph{Initial conditions} At $t = 0$, $K$ geodesic disks (spherical caps) are placed on the unit sphere with centres $\hat{\vec{c}}_k$ drawn uniformly via the $(z, \varphi)$ parameterisation, angular radii $r_k \in [0.3, 1.0]$ rad ($\approx 17^\circ$ to $57^\circ$) drawn uniformly, and piecewise-constant primitive states $(h_k, u_k, v_k)$ with $h_k \in [0.5, 2.0]$ and $u_k, v_k \in [-0.5\delta, 0.5\delta]$. The flow-strength parameter $\delta \in [0, 1]$ scales velocities only; depth jumps are independent of $\delta$, so even at $\delta = 0$ the initial pressure imbalance drives full-amplitude Riemann fans. A separate background state $(h_\textup{bg}, u_\textup{bg}, v_\textup{bg})$ fills any region not claimed by a cap. Cells in cap overlaps are assigned by a painter's algorithm in cap index order. At IC time every solver cell is sub-sampled $4 \times 4$ in computational coordinates and the cell IC is the mean of the primitives across the $16$ sub-points, with momentum projected to 3-D Cartesian at the cell centre; this eliminates staircase artefacts and gives the solver a well-resolved initial shock width of $\sim 1$ FV cell. Because cap centres are drawn uniformly on $S^2$ from the outset, no separate $\mathrm{SO}(3)$ tilt step is required, shock interfaces are already non-aligned with any computational axis. The $\delta$-scaled velocity bounds keep the initial flow strictly subcritical at all $\delta$, with $\max\,\textup{Fr} \approx 0.23\delta$, peaking at $\approx 0.23$ when $\delta = 1$.

\paragraph{Parameter grid} Varying parameters are the number of caps $K$ and the velocity scaling $\delta$, plus the IC seed. For each $(K, \delta)$ block the seed dictates all stochastic quantities (cap centres, radii, per-cap states, background state); different $(K, \delta)$ blocks at the same seed draw their caps independently, so there is no subset relationship between cap counts. Detailed values are in \autoref{tab:datasetparams}.

\subsection{\heldsuareztt}
\begin{figure}[t]
    \centering
    \includegraphics[width=\linewidth]{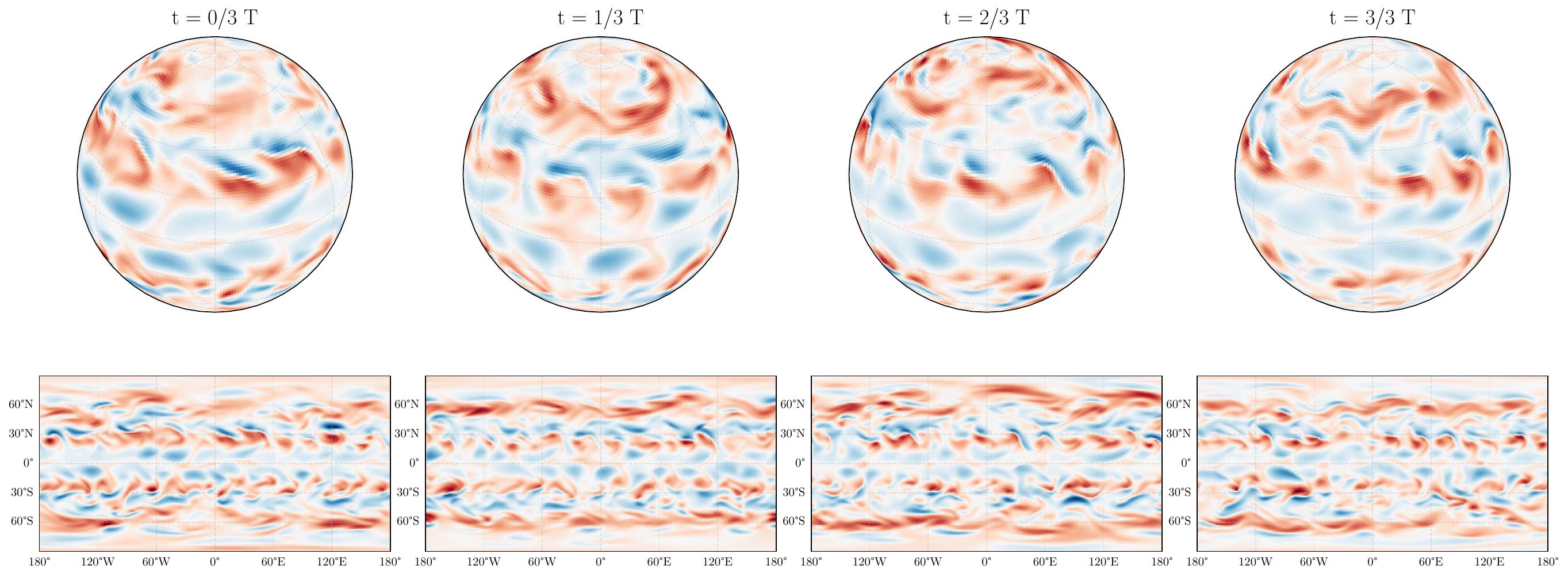}
    \caption{Evolution of the zonal wind at \qty{1000}{\hecto\pascal} (the lowest model level) along a \heldsuareztt trajectory. Top row: states depicted on the sphere. Bottom row: states depicted on a latlon grid.
    Color scale limits (vmin/vmax) are shared.} 
    \label{fig:held-suarez-4-evolution}
\end{figure}
\paragraph{Underlying physics} The benchmark is the dry Held--Suarez idealised general circulation model~\cite{held1994proposal}, a primitive-equation atmospheric setup designed to exercise the dynamical core of a GCM in the absence of moisture, radiation, or topography. The model evolves horizontal winds and temperature under adiabatic dynamics, Newtonian relaxation toward a prescribed equilibrium temperature $T_\textup{eq}(\phi, p)$, and Rayleigh friction concentrated in the lower atmosphere. The forcing is parameterised by three rates: a free-atmosphere Newtonian-cooling rate $k_a$, a stronger near-surface cooling rate $k_s$, and a lower-level Rayleigh drag rate $k_f$, each set by the standard Held--Suarez formulas. The lower boundary is flat and homogeneous (no topography, no surface fluxes); the atmosphere is dry, with no moisture, radiation, or orography. The results are jet structures, with eddies and storm tracks generated by baroclinic instability of the equilibrium gradient. Unlike the original benchmark by \citet{held1994proposal}, the ClimaAtmos.jl solver uses a nonhydrostatic setting.

\paragraph{Simulation details} Each trajectory is integrated with ClimaAtmos.jl, the atmospheric component of the CliMA software stack~\cite{yatunin2026climatemodeling}, on a cubed-sphere grid. Each cube surface uses a Gauss--Lobatto--Legendre (GLL) grid tiled with $24 \times 24$ elements per face, with a degree-3 polynomial in each element. Additionally, the solver uses 31 vertical levels up to $60000\si{\meter}$. We use the built-in export to NetCDF which automatically re-grids the cubed-sphere to a $144\times288$ lat-lon grid and re-maps the elevation-based vertical levels to 37 pressure-based levels. We take 8 of the 13 usual ERA5 pressure levels used in ML weather forecasting~\cite{rasp2024weatherbench2} (50, 100, 250, 500, 700, 850, 925, 1000 $\si{\hecto\pascal}$).

Each production runs for 565 days, with the first $200$ days discarded as spin-up and the subsequent 365 days retained as 6-hourly output. This
  discards the seeded transient and preserves only the statistically meaningful post-spin-up evolution. GPU-accelerated generation (using either NVIDIA RTX 4090 or NVIDIA L40S) results in a per-run wall-time on the order of six hours %
.

\paragraph{Fields} Available fields are the zonal velocity $u$, the meridional velocity $v$, and the air temperature $T$ at 50, 100, 250, 500, 700, 850, 925, 1000 $\si{\hecto\pascal}$, plus surface pressure $p_s$ (cf.~\autoref{tab:datasetsrecap}).

\paragraph{Initial conditions} Each run is initialised using a decaying temperature profile from a surface temperature of $\SI{290}{\kelvin}$ to $\SI{220}{\kelvin}$ using a $\tanh(z/H_t)$-decay schedule with $H_t=\SI{8}{\kilo\meter}$. To this initial temperature profile we add a small seeded random perturbation, a random field of amplitude $\SI{0.1}{\kelvin}$ but only for $z< \SI{5}{\kilo\meter}$.

\paragraph{Parameter grid} Varying parameters are the angular velocity of the planet $\Omega$, given as a factor of the Earth's rotation $\Omega_{\text{earth}} \approx \SI{7.292e-5}{\per\second}$, the difference between the equator and pole equilibrium temperatures $\Delta T_y$ (the coefficient of the $\sin^2\phi$ term in $T_\textup{eq}$) and the vertical potential-temperature gradient $\Delta\theta_z$, which sets the static stability of the equilibrium state by controlling how strongly the target potential temperature increases with height (it is the coefficient of the $\log(p/p_0)\cos^2\phi$ term in $T_\textup{eq}$).

This parameter space is intentionally centered on the canonical Held--Suarez setting.

Detailed values are in \autoref{tab:datasetparams}. The Held--Suarez benchmark keeps the rotation axis aligned with the computational pole because that is the convention assumed by the equilibrium profile $T_\textup{eq}(\phi, p)$ and by the meridional structure of the forcing.

\paragraph{References} \cite{held1994proposal,yatunin2026climatemodeling}

\subsection{\oceantt}
\begin{figure}[t]
    \centering
    \includegraphics[width=\linewidth]{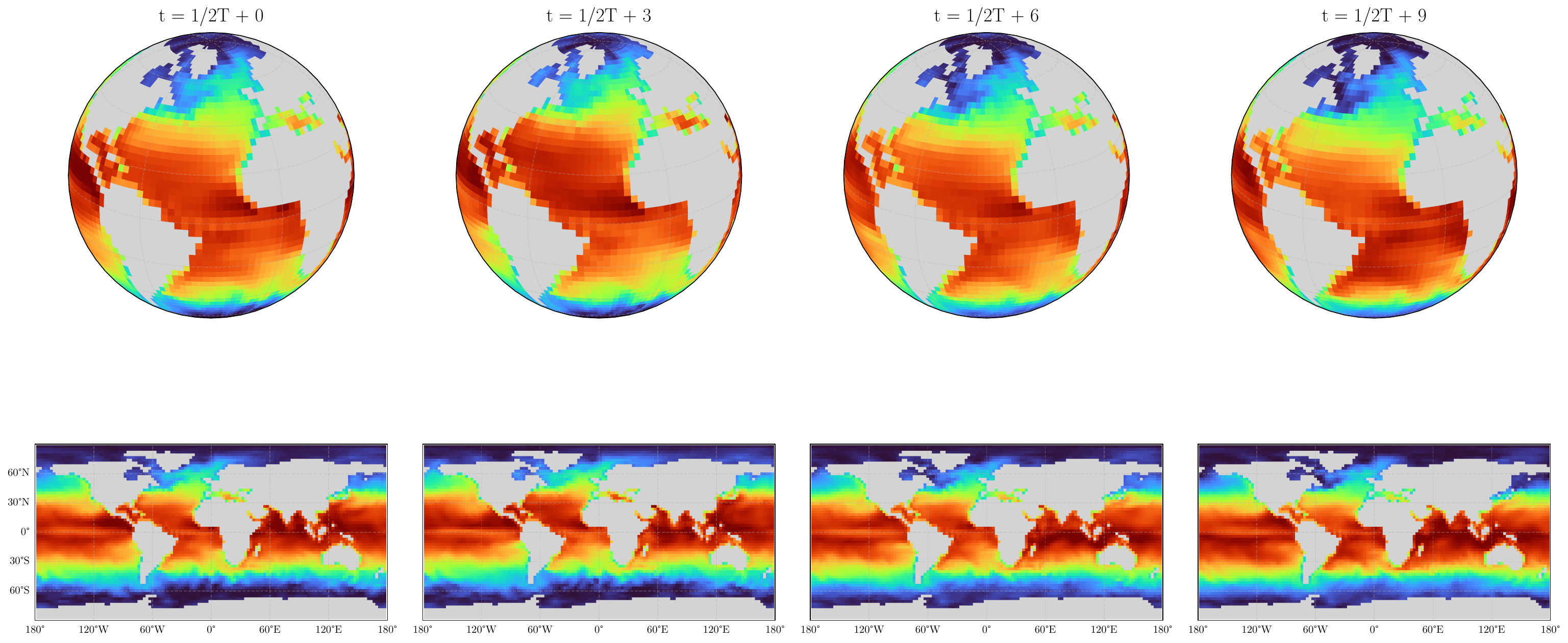}
    \caption{Evolution of the Temperature variable of an \oceantt trajectory, selected to show the seasonality by plotting a snapshot every three months (most clearly visible in the temperature of the Mediterranean sea). Top row: states depicted on the sphere. Bottom row: states depicted on a latlon grid. Color scale limits (vmin/vmax) are shared.}
    \label{fig:ocean-dynamics-4-seasons}
\end{figure}

\paragraph{Underlying physics} The benchmark is a primitive-equation ocean simulation configured from the MITgcm \texttt{global\_ocean.cs32x15} verification case~\cite{marshall1997finite,adcroft2004implementation}. It features realistic bathymetry, Levitus hydrographic initial conditions and monthly surface restoring~\cite{levitus1982climatological}, Trenberth wind-stress climatology~\cite{trenberth1989global}, shi/ncep heat and freshwater fluxes, GM/Redi mesoscale-eddy parameterisation~\cite{gent1990isopycnal}, and GGL90 vertical mixing. The ocean is non-autonomous: both the surface restoring fields and the atmospheric fluxes are subject to annual cycles, meaning the same instantaneous state can map to different $30$-day snapshots, depending on the hidden seasonal phase. 

\paragraph{Simulation details} Each trajectory runs on a cubed-sphere grid with six $32\times32$ faces and $15$ depth levels (depths at cell centers range from $\SI{25}{\meter}$ to $\SI{4855}{\meter}$). The simulation includes a non-linear free surface (representing a real fresh-water flux) and vector-invariant momentum. 
Runs have a discarded warm-up initialization of roughly $200$ model years, and start from the \texttt{pickup.0000072000} file (shipped with the tutorial). The production run then consists of $100$ years ($\num{36000}$ steps with $\Delta t = 1\  \text{day}$). Snapshots are written every $30$ simulated days, with the first $12$ dropped (accounting for $\approx 1$ model year of parameter-change adjustment transient). Each finalised trajectory therefore contains $1189$ snapshots covering $\sim 99$ model years (at $360$ days each); the time coordinate is shifted so that $t[0] = 0$. Native solver outputs are stored face-major as $(\textup{time}, \textup{field}, 6, 32, 32)$ and rotated/regridded to a $64 \times 128$ lat-lon grid at training time. Vector rotation from face-aligned to geographic east/north uses the static $(\cos\alpha, \sin\alpha)$ angles supplied by MITgcm. Per-run wall is $\sim25$--$30$ minutes single-CPU for a full $100$-year trajectory.

\paragraph{Fields} Available fields are the potential temperature and salinity at all tracer levels $k = 1, \dots, 15$, the face-aligned velocities $(u, v)$ at the corresponding velocity levels (rotated to geographic east/north before regridding), and the sea-surface height $\eta$ (cf.~\autoref{tab:datasetsrecap}). Each dynamic field carries a level-dependent ocean mask $\textup{mask}_k$, $k=1,\ldots,15$, while the velocity fields use the corresponding masks defined on the staggered velocity faces. The sea-surface height $\eta$ has a separate column mask $\textup{mask}_\eta$, which marks any column with at least one wet level. Land cells are stored as $0.0$. For the purpose of calculating normalization statistics and whenever we calculate a loss or metric, we mask accordingly, so that the model is scored only on real ocean.

\paragraph{Initial and boundary conditions} The cs32x15 verification case provides realistic bathymetry, Levitus 3-D initial $T$ and $S$, Levitus monthly surface restoring, Trenberth wind stress, and shi/ncep heat and freshwater fluxes. %

\paragraph{Parameter grid} Varying parameters are the GM/Redi background diffusivity $\kappa_\textup{GM}$, the horizontal viscosity $A_h$, the vertical diffusivity $K_v$, and the surface restoring timescales for temperature ($\tau_T$) and salinity ($\tau_S$). The current grid is a $3^5 = 243$ tensor product designed to span weak, standard, and strong mixing regimes; detailed values are in \autoref{tab:datasetparams}. The original $A_h$ grid included $A_h = \SI{e6}{\meter\squared\per\second}$, leading to the model crossing \texttt{CFL=1}. The grid was then tightened to $(1.5\times 10^5,\, 3\times 10^5,\, 5\times 10^5)\,\textup{m}^2/\textup{s}$, all below the unstable boundary, and a $365$-step preflight on the new grid showed $\max\,\text{advcfl}_\text{wvel} \approx 0.094$.

\paragraph{References} \cite{marshall1997finite,adcroft2004implementation}

\section{Models}
\subsection{Flower}
\label{app:flower}
The Flower architecture~\cite{muser2026flowerswarpdriveneural} is a U-Net whose core mixing primitive is a multi-head warping block called \emph{SelfWarp}. Given an input feature map $u \in \mathbb{R}^{C \times H \times W}$, a SelfWarp layer produces $H$ independent (flow, value) heads via two parallel $1\times1$ convolutions: a flow head $\mathbf{F}_\text{flow}: u \mapsto \delta \in \mathbb{R}^{H \times 2 \times H' \times W'}$ predicting per-pixel displacements, and a value head $\mathbf{F}_\text{val}: u \mapsto v \in \mathbb{R}^{H \times C/H \times H' \times W'}$. For each head $h$, the output is sampled bilinearly at displaced coordinates,
\begin{equation}
    u_\text{warp}^{(h)}(x) = v^{(h)}\bigl(x + \delta^{(h)}(x)\bigr),
\end{equation}
implemented through PyTorch's \texttt{grid\_sample} with a custom dispatch that supports per-axis periodic, zero, or border padding. The head outputs are concatenated and a residual connection plus normalization (GroupNorm) and a GELU activation close the block. Each block thus learns a content-adaptive, multi-scale transport operator: rather than mixing neighbours through a fixed kernel, it samples wherever the displacement field points.

These SelfWarp blocks are arranged in a U-Net with $L$ levels (default $L=4$), where the channel width doubles per level via $\mathrm{lifting\_dim} \cdot 2^i$. The encoder downsamples with strided $2 \times 2$ convolutions and the decoder upsamples with the matching transposed convolutions; a single bottleneck SelfWarp block sits at the deepest level and skip connections concatenate encoder activations into the decoder path. Inputs are augmented with a normalized coordinate grid before lifting, and an optional FiLM pathway broadcasts a metadata vector into every block's normalization layer to support conditional rollouts. The default 2D configuration uses lifting dim $160$, $40$ heads, $40$ groups, and zero / periodic boundaries along latitude / longitude.

Flower is highly effective on flat 2D and 3D PDE benchmarks but inherits two structural assumptions that break on $S^2$: (i) displacements live in pixel coordinates, so a fixed $\delta$ corresponds to wildly different physical arc lengths near the equator versus near the poles; and (ii) the strided downsampler is a vanilla \texttt{Conv2d}, whose square receptive field severely distorts under the lat--lon metric. \netname~(\S\ref{app:model}) replaces both primitives while preserving the multi-head warp-and-residual structure.

\subsection{\netname}
\label{app:model}
\netname keeps Flower's U-Net skeleton and head-based mixing but swaps every grid-sensitive primitive for a natively spherical analogue. Two changes are central: (i) the displacement prediction is reframed as a tangent-plane vector, projected onto the sphere via Rodrigues' formula; and (ii) all spatial pooling is performed in the spherical-harmonic domain rather than via strided convolution. Feature maps stay on equiangular nodes throughout.

\paragraph{Tangent SelfWarp} The flow head predicts a 2-vector $(du^{(h)}, dv^{(h)})$ at every grid point and head, interpreted as a displacement in the local tangent plane spanned by the unit east and north vectors $\mathbf{e}_E$ and $\mathbf{e}_N$. Writing the base-point unit vector as $\mathbf{p} \in S^2$, we form the tangent displacement $\mathbf{d}_{3D} = du \, \mathbf{e}_E + dv \, \mathbf{e}_N$ and apply the closed-form Rodrigues exponential map,
\begin{equation}
    \mathbf{p}' = \mathbf{p} \cos\theta + \mathbf{d}_{3D} \frac{\sin\theta}{\theta},
    \qquad \theta = \lVert \mathbf{d}_{3D} \rVert = \sqrt{du^2 + dv^2},
\end{equation}
which transports $\mathbf{p}$ along the great circle defined by the tangent direction by an arc length $\theta$. The new point $\mathbf{p}'$ is converted back to spherical coordinates $(\lambda', \varphi')$, then to grid-sample coordinates via $g_x = \lambda'/\pi - 1$ for longitude (analytic, since longitude rows are equispaced) and via a monotonic ascending-lat lookup $\varphi' \mapsto g_y$ for latitude (which on equiangular grids reduces to an affine formula and on Legendre--Gauss grids is the exact per-row coordinate). The value head is sampled at this great-circle-displaced location with periodic-longitude / border-latitude padding. Because the construction is purely 3D and the lookup is monotonic, displacements that cross a pole are handled naturally: no special longitude wrap, no polar reflection.

\paragraph{Spectral coarsening} Each U-Net resolution change is implemented through a Spherical Harmonic Transform pair on equiangular nodes. The downsampler $\mathrm{Spectral2xDown}: (H_\text{hi}, W_\text{hi}) \to (H_\text{lo}, W_\text{lo})$ projects the input onto real spherical harmonics with degree $\ell_\text{max} = H_\text{lo}$ and order $m_\text{max} = W_\text{lo}/2 + 1$, then inverse-transforms onto the coarse grid; a single $1\times1$ convolution and a GELU adjust the channel count from $C_\text{in}$ to $C_\text{out}$, with no normalization around the SHT pair. The upsampler reverses this: a $1\times1$ convolution to expand channels, an SHT on the low-resolution grid, and an inverse SHT onto the fine grid. With $\ell_\text{max}, m_\text{max}$ set by the coarse resolution, the round-trip is information-preserving (low-pass). The architecture is therefore convolution-free in the spatial-mixing path: only $1\times1$ pointwise convolutions remain, which never see the lat--lon metric.

\paragraph{Internal grid} \netname runs on the equiangular grid at every level. The SHT is not quadrature-exact on equiangular nodes, but the residual aliasing is small compared to the explicit low-pass at each resolution change, and staying on a single grid saves four boundary SHTs per forward pass.

\paragraph{Conditioning and lifting} Inputs are concatenated with a 3D Cartesian unit-sphere embedding $(x, y, z)$ --- three channels rather than the two raw $(\lambda, \varphi)$ coordinates — before lifting via a $1\times1$ convolution. This avoids the discontinuity at $\lambda = 0/2\pi$ and gives the network direct access to a coordinate frame in which great-circle distance is linear. Each FlowerBlock follows the pre-norm pattern $x \mapsto x + W \cdot \mathrm{warp}(\mathrm{norm}(x))$, with GELU and optional dropout.

\paragraph{2.5D version}\label{app:model-2p5d} Treating the 3D datasets as 2D by flattening vertical levels into channels is a crude treatment, but adapting neural operators to 3D usually incurs significant increase in computational complexity. We design a simple extension through 3D by interspersing the 2D warps with separate vertical mixing layers. For our model, we simply stack channels vertically and then use a $ 1\times1 $ convolution on the resulting field to increase/decrease the number of channels, approximately halving at each stage (for \heldsuareztt, we use $ [9, 5, 3, 2] $ vertical layers, i.e. a change to $n/2+1$ with each downsampling step, for \oceantt $[7,4,2,1]$). In the downsampling path, we situate this operation before the horizontally-acting SHT downsampling, in the upsampling path before. This is the most computationally efficient order.

For \heldsuareztt we use two different encoders, one for the surface pressure and one shared across all atmospheric layers. The results are concatenated, giving $1+8=9$ levels. We decode in a similar manner (using two decoders) back to separate surface pressure and atmospheric variables.

For \oceantt, we similarly use an encoder for the surface variable and a separate encoder shared across vertical layers.

\subsection{Fourier Neural Operator and Spherical Fourier Neural Operator}
Fourier Neural Operators~\cite{li2021fourierneuraloperatorparametric} (FNOs) perform spatial mixing as global convolutions in the spectral domain. The input is lifted pointwise from $C_\text{in}$ to a hidden width $d$ and passed through a stack of Fourier layers. A Fourier layer takes the FFT of its input, multiplies a low-frequency truncation of the coefficients by a learned complex tensor, and applies the inverse FFT; this is a global convolution with a band-limited kernel. A parallel pointwise convolution carries the high-frequency residual. The two branches are summed and passed through a GELU. A final pointwise projection returns the hidden state to $C_\text{out}$ output channels. Because the spatial mixing uses an FFT, the architecture assumes a flat, doubly-periodic grid: a poor fit to lat--lon data, where only longitude is periodic and the metric varies with latitude.

Spherical Fourier Neural Operators (SFNOs)~\cite{bonev2023sphericalfourierneuraloperators} replace the FFT with a real Spherical Harmonic Transform (SHT). A truncated harmonic expansion takes the place of the band-limited Fourier kernel, and an inverse SHT returns to physical space; the residual pointwise convolution, MLP, and normalization are unchanged. SFNO is also the spectral backbone of FourCastNet v2 and v3~\cite{bonev2023sphericalfourierneuraloperators,bonev2025fourcastnet3geometricapproach}.

We instantiate both models from \texttt{torch\_harmonics.examples.models}, with FNO obtained from the same scaffold by replacing the SHT pair with an \texttt{rfft2}/\texttt{irfft2} pair on the periodic-padded grid. Hyperparameters are held fixed across the two: embedding dimension $192$, four Fourier layers, MLP ratio $2$, hard-thresholding fraction $0.5$, instance normalization, GELU activation, no positional embedding, and direct (non-residual) prediction. The spectral truncation is set as a fraction of the grid resolution, so parameter counts are grid-independent.

\subsection{$\mathbb{R}^2$ and $\mathcal{S}^2$ Transformer}
\label{app:transformers}
The transformer architecture~\cite{vaswani2017attention} mixes information through scaled dot-product attention. Vision Transformers~\cite{dosovitskiy2021imageworth16x16words} adapt the primitive to images: partition the input into non-overlapping patches, lift each patch to a token via a learned linear projection, and apply global self-attention across the resulting sequence with additive positional embeddings to break attention's permutation invariance. Two parts of this recipe transfer poorly to the sphere. A square patch tokenizer distorts under the lat--lon metric, the same problem that breaks Flower's \texttt{Conv2d} downsampler (\S\ref{app:flower}). Additive Cartesian positional embeddings, in turn, carry no notion of geodesic distance.

Both transformer baselines we use come from \citet{bonev2025attentionsphere} and differ in only two respects: the tokenizer/detokenizer (a standard convolution vs.\ DISCO) and the attention neighborhood (a Cartesian window on the lat--lon grid vs.\ a geodesic ball on $S^2$). \emph{LocalRTransformer} keeps a flat lat--lon grid but replaces dense attention with NATTEN's \texttt{NeighborhoodAttention2D}: each token attends only to a $7\times7$ rectangular window, so per-layer cost is linear in the sequence length and the receptive field grows with depth. \emph{LocalS2Transformer} replaces the rectangular $3\times3$ tokenizer with a DISCO encoder/decoder pair --- a discrete-continuous spherical convolution with a $5\times4$ piecewise-linear filter basis --- and the $7\times7$ window with \texttt{NeighborhoodAttentionS2}, whose attention windows are geodesic balls on $S^2$. We follow the configurations of~\citet{bonev2025attentionsphere}: both models use spectral positional embeddings, an encoder downsampling factor of $2$, embedding dimension $128$, $4$ blocks, a single attention head (head dim $128$, a power of two as NATTEN's flex-attention backend requires on the planar variant), MLP ratio $2$, instance normalization, and GELU activations. Width and depth are matched, so the two models are parameter-equivalent at $\sim$545K trainable parameters on \texttt{planetswe}. %

Spherical physics models built on transformers include Pangu-Weather~\cite{bi2022panguweather3dhighresolutionmodel}, Stormer~\cite{nguyen2024scaling}, and Aurora~\cite{bodnar2024aurora}.

\subsection{GAOT}
\label{app:gaot}
GAOT~\cite{wen2026geometryawareoperatortransformer} is an attention-based model designed to work on arbitrary domains; see~\cite{wen2026geometryawareoperatortransformer} for details. We use the authors' implementation with the hyperparameters of their time-dependent example configuration (\texttt{ns\_gauss}): a $64\times64$ latent token grid with patch size $2$, a $3$-layer transformer of width $256$, and multiscale graph encoder and decoder with hidden size $64$ and radius $0.033$. The lat--lon grid is treated as a plane ($2$D coordinates). The resulting model has $\sim$3.4M parameters.

\section{Experiment Details}
\label{app:experiment}

This appendix collects the training and evaluation protocol used for every
number in \S\ref{sec:experiments}. Dataset specifications (resolutions,
channel counts, trajectory layouts, simulator settings) are in
\S\ref{app:dataset}; model architectures and per-model hyperparameters
(hidden widths, depths, parameter counts) are in \S\ref{app:flower} and
\S\ref{app:model}.

\subsection{Task formulation and splits}

Given four consecutive frames $\{x_{t-3}, x_{t-2}, x_{t-1}, x_t\}$, the
model predicts $\hat{x}_{t+1}$. Frames are stacked along the channel
dimension before the lifting convolution, so the spatial backbone sees a
single $(C \cdot 4, H, W)$ tensor. The head outputs one frame, and rollout
slides the four-frame buffer forward in time.

For the results published in this paper, the Z-axis of any 3D dataset is simply added to the channel axis.
Each dataset is split $80/10/10$ into train, validation, and test trajectories. Splits are at the trajectory level (no frame leaks across splits) and are deterministic given the dataset seed; PlanetSWE inherits The Well's published split unchanged. For \galewskytt we split along physical trajectory, grouping different rotations of the same trajectory together.

Inputs and targets are z-score normalized per channel using statistics computed on the training split, with the statistics inlined in each \texttt{configs/data/*.yaml}. The same statistics denormalize predictions before metric evaluation, so reported errors are in the original physical units.

\subsection{Two-phase training}

Each model is trained for $25$ epochs in two phases:

\begin{itemize}
    \item \textbf{Phase~1 (single-step), $20$ epochs.} The loss is computed on a single one-step prediction $\hat{x}_{t+1}$ against the ground truth $x_{t+1}$.
    \item \textbf{Phase~2 (autoregressive), $5$ epochs.} The model is unrolled for two steps: $\hat{x}_{t+1}$ is fed back into the four-frame buffer and the model produces $\hat{x}_{t+2}$. The loss averages the latitude-weighted MSE over both steps, $\mathcal{L}_{\text{AR}} = \tfrac{1}{2}\bigl[\mathcal{L}(\hat x_{t+1}, x_{t+1}) + \mathcal{L}(\hat x_{t+2}, x_{t+2})\bigr]$, with gradients flowing through both unrolls (no teacher forcing on step~2).
\end{itemize}

The warmup--cosine schedule (below) covers the full $25$ epochs across both phases; only the loss formulation changes at the phase boundary. Validation runs every epoch in both phases: a single-step pass mirroring the Phase~1 loss and a full $20$-step rollout mirroring the test-time setting in \autoref{tab:results41}. We do not reload a best-validation checkpoint before testing; the test row uses the model state at the end of training, so val and test numbers reflect the same parameters.

\subsection{Loss function}

The training loss is a cosine-of-latitude-weighted MSE,
\begin{equation}
\mathcal{L}(\hat y, y)
\;=\;
\frac{1}{B\,C\,H\,W}\sum_{b,c,h,w} w_h \,\bigl(\hat y_{b,c,h,w}-y_{b,c,h,w}\bigr)^2,
\qquad
w_h \;=\; \frac{\cos(\mathrm{lat}_h)}{\frac{1}{H}\sum_{h'}\cos(\mathrm{lat}_{h'})}.
\label{eq:lat-weighted-mse}
\end{equation}
Equiangular grids oversample near the poles, so an unweighted MSE lets high-latitude pixels dominate the gradient; the $\cos(\mathrm{lat})$ factor turns the discrete sum into a Monte-Carlo estimate of the spherical $L^2$ error so each unit of solid angle contributes equally. The weights are normalized to mean~$1$ along the latitude axis, which keeps $\mathcal{L}$ on the same numerical scale as plain MSE and lets learning-rate and
gradient-clipping settings transfer between the weighted and unweighted regimes. The same convention is used in
GraphCast~\cite{lam2023graphcast} and FourCastNet~\cite{pathak2022fourcastnetglobaldatadrivenhighresolution}.
For masked-loss datasets (the global-ocean variants, where land cells carry no signal), the squared error is multiplied by a per-channel \texttt{valid\_mask} and the denominator is the weighted count of valid cells. There are no auxiliary terms (no spectral loss, no divergence penalty, no consistency or pushforward regulariser, no per-channel reweighting).

For validation, we use latitude-weighted (i.e. area-weighted) VRMSE, defined for a single field as
\begin{equation}
    \operatorname{VRMSE}(y, \hat{y}) = \sqrt{\frac{\mathbb{E}_d\left[w_h (y-\hat{y})^2\right]}{\operatorname{Var}_w(y) + \epsilon}},
    \qquad
    \operatorname{Var}_w(y) = \mathbb{E}_d\left[w_h\bigl(y - \mathbb{E}_d[w_h y]\bigr)^2\right],
    \label{eq:vrmse}
\end{equation}
where $w_h$ is the same $\cos(\mathrm{lat})$ weight used in the training loss (Eq.~\ref{eq:lat-weighted-mse}), and $\operatorname{Var}_w(y)$ is the latitude-weighted variance of the ground truth over spatial axes, using the same weighted mean and second moment. We set $\epsilon = 10^{-7}$. VRMSE is computed per sample and per channel, then averaged over the batch, and finally averaged over channels to obtain a single scalar; because the ratio is taken before averaging, this differs in general from the ratio of batch-averaged MSE to batch-averaged variance.

\subsection{Optimizer and schedule}

We optimize with AdamW~\cite{loshchilov2019decoupledweightdecayregularization} at base learning rate (see the next paragraph), weight decay $1 \times 10^{-4}$, default betas $(0.9, 0.999)$, and $\epsilon = 10^{-8}$. The schedule (\texttt{fots.utils.build\_warmup\_cosine}) is a one-epoch warmup at $10^{-3}$ of the base rate, with subsequent cosine annealing from base-rate over the remaining $24$ epochs down to $\eta_{\min} = 10^{-5}$. The schedule steps once per epoch.
Gradients are globally clipped at $L_2$-norm $1.0$. We use no EMA and no SWA.

For each (model, dataset) cell of \autoref{tab:results41} we sweep the base learning rate over $\{10^{-4},\, 5\times 10^{-4},\, 10^{-3}\}$ and select the run with the lowest single-step validation loss at the end of training.

\subsection{Hardware and distributed setup}

Runs were generally trained on nodes with four NVIDIA GH200 Grace--Hopper Superchips ($96$~GiB HBM3 each) on the \texttt{normal} partition. A training job occupies one node and runs across the four GPUs via PyTorch \texttt{DistributedDataParallel}. Some of the \texttt{planetswe} trainings were done instead on a single NVIDIA H200.

All training runs were set up with a time-limit of 24 hours due to a limited compute budget. Most models finish their training well before the limit, the only exception are the transformer models on some of the larger datasets.

\paragraph{Per-(model, resolution) batch sizes} Activation memory scales linearly in batch size, so we tune the per-GPU batch per architecture from a memory-probe sweep on H200 (\texttt{scripts/probe\_memory*.sbatch}) and
scale the H200 maximum down to GH200 by the HBM ratio $96/140 \approx 0.686$.

For the LocalS$^2$Transformer, we did not manage to finish training on a number of higher-resolution datasets within a reasonable timeframe.

\subsection{Evaluation}

The single-step and rollout numbers in \autoref{tab:results41} are
computed on the held-out test split with a number of area-weighted spherical metrics.
All spatial means use the same $\cos(\mathrm{lat})$ weights as the training loss. The rollout column unrolls the model autoregressively for $\min(20,\, T_{\text{out}})$ steps from each test trajectory's first four-frame window and aggregates per-step metrics into a single average; per-step curves are logged but not included in the main table.

\subsection{Ablations}
\label{app:ablation}

We separately ablate the proposed spherical primitives. We run experiments replacing the tangent warp with lat--lon warping, the SHT coarsening with pooling and the $(x,y,z)$ embedding with simple raw lat--lon embeddings.
Each variant is trained and selected with the same protocol as in \autoref{tab:results41}. Since ablating all three primitives yields a Flower-like architecture, we consider the Flower2D row of \autoref{tab:ablation} as an everything-ablated variant.

\begin{table}[h]
  \centering
  \caption{Leave-one-out ablation of \netname: area-weighted VRMSE for single-step prediction and 1:20 rollout. Each row replaces one spherical primitive with its flat counterpart. \netname and Flower2D rows are from \autoref{tab:results41}. Best per column in bold.}
  \label{tab:ablation}
  \small
  \begin{tabular}{ll cccc}
    \toprule
    & & \galewskytt & \mickelintt & \shockcapstt & \heldsuareztt \\
    \midrule
    \multirow{5}{*}{\rotatebox[origin=c]{90}{\textit{Next-step}}}
      & \netname       & 0.0037 & 0.0441 & 0.0082 & 0.0274 \\
      & \quad w/ lat--lon warp  & \textbf{0.0015} & 0.0486 & 0.0080 & \textbf{0.0243} \\
      & \quad w/ pooling        & 0.0028 & 0.0469 & 0.0118 & 0.0282 \\
      & \quad w/ lat--lon emb.  & 0.0028 & \textbf{0.0434} & \textbf{0.0079} & 0.0272 \\
      & Flower2D       & 0.0055 & 0.0700 & 0.0160 & 0.0290 \\
    \midrule
    \multirow{5}{*}{\rotatebox[origin=c]{90}{\textit{1:20 Rollout}}}
      & \netname       & 0.0422 & 0.8915 & 0.0414 & 0.2067 \\
      & \quad w/ lat--lon warp  & \textbf{0.0327} & 0.8983 & 0.0403 & \textbf{0.1956} \\
      & \quad w/ pooling        & 0.0466 & 0.9067 & 0.0527 & 0.2275 \\
      & \quad w/ lat--lon emb.  & 0.0370 & \textbf{0.8795} & \textbf{0.0399} & 0.2029 \\
      & Flower2D       & 0.0588 & 0.9583 & 0.0635 & 0.2214 \\
    \bottomrule
  \end{tabular}
\end{table}

We find that \netname's performance improvement is cumulative rather than driven by any single primitive. In many cases, removing one primitive changes the result only by a little (on \galewskytt the next-step error with lat--lon warping is actually lower than the tangent warp).
Removing all three, however, generally results in a significant performance degradation (\autoref{tab:results41}).

 On \heldsuareztt all variants, including Flower2D, are close to the full model. We assume that this is because the dynamics here are largely aligned with latitude lines, which is when spherical adaptation matters least.
\subsection{Bootstrap confidence intervals}
\label{app:bootstrap}
We quantify differences between the test set performance of models via a bootstrap over test trajectories on \texttt{planetswe} (12 test trajectories) and \mickelintt (76).
We re-evaluate every model and record its VRMSE per test trajectory: the mean over all next-step windows of that trajectory, and the mean over the 20 rollout steps. We then draw $10^4$ resamples of the test trajectories with replacement and recompute each metric on every resample, exactly as in \autoref{tab:results41} (next-step: mean over all windows of the drawn trajectories; rollout: mean over the drawn trajectories). All models are scored on the same resamples and we calculate the ratio of a model's error to that of \netname in a paired comparison. We report 95\% percentile intervals.

\autoref{tab:bootstrap} shows that every ratio interval lies above one, i.e.\ \netname has the lowest error on every resample, against every baseline, on both datasets and both metrics.

\begin{table}[htbp]
  \centering
  \footnotesize
  \setlength{\tabcolsep}{2.5pt}
  \caption{Bootstrap over test trajectories ($10^4$ resamples, 95\% percentile intervals in brackets). Last row: VRMSE of \netname. Other rows: ratio of each model's VRMSE to that of \netname on the same resample (above $1$: \netname is better).}
  \label{tab:bootstrap}
  \begin{tabular}{@{}l cc cc@{}}
    \toprule
    & \multicolumn{2}{c}{\texttt{planetswe}} & \multicolumn{2}{c}{\mickelintt} \\
    \cmidrule(lr){2-3} \cmidrule(l){4-5}
    Model & \textit{Next-step} & \textit{1:20 Rollout} & \textit{Next-step} & \textit{1:20 Rollout} \\
    \midrule
    \textsc{FNO}          & 23.7 {\scriptsize[23.1, 24.5]} & 18.9 {\scriptsize[15.1, 23.1]} & 13.2 {\scriptsize[12.6, 13.9]} & 1.19 {\scriptsize[1.15, 1.24]} \\
    \textsc{SFNO}         & 15.9 {\scriptsize[14.9, 17.0]} & 8.23 {\scriptsize[7.75, 8.77]}    & 9.37 {\scriptsize[8.73, 9.96]}    & 1.30 {\scriptsize[1.25, 1.36]} \\
    \textsc{GAOT}         & 73.0 {\scriptsize[67.9, 78.3]} & 8.47 {\scriptsize[8.00, 8.98]}    & 7.17 {\scriptsize[6.95, 7.37]}    & 1.07 {\scriptsize[1.04, 1.12]} \\
    \textsc{LR$^2$Trnsf.} & 10.2 {\scriptsize[9.61, 10.9]}  & 7.33 {\scriptsize[6.49, 8.28]}    & 6.31 {\scriptsize[5.92, 6.72]}    & 1.30 {\scriptsize[1.23, 1.37]} \\
    \textsc{LS$^2$Trnsf.} & 7.50 {\scriptsize[6.94, 8.11]}    & 52.3 {\scriptsize[45.6, 60.2]} & 4.67 {\scriptsize[4.34, 5.01]}    & 1.11 {\scriptsize[1.07, 1.15]} \\
    \textsc{Flower2D}     & 1.51 {\scriptsize[1.48, 1.53]}    & 1.59 {\scriptsize[1.52, 1.68]}    & 1.59 {\scriptsize[1.53, 1.66]}    & 1.07 {\scriptsize[1.04, 1.11]} \\
    \midrule
    \textsc{\netname} & 0.0005 {\scriptsize[0.0005, 0.0006]} & 0.0107 {\scriptsize[0.0093, 0.0122]} & 0.0440 {\scriptsize[0.0403, 0.0479]} & 0.892 {\scriptsize[0.861, 0.924]} \\
    \bottomrule
  \end{tabular}
\end{table}

\subsection{Computational cost}
\label{app:cost}
To measure the compute cost of \netname and see where it comes from, we time a single training step (forward and backward pass, no optimizer step) on one NVIDIA H200, with synthetic inputs and each model's training batch size, so that any additional overhead such as data loading does not pollute the comparison. We additionally time every component in isolation (forward and backward on the inputs it receives during the step) and sum over its calls. \autoref{tab:cost_breakdown} gives the result. Per sample, \netname costs $2.5\times$ as much as Flower2D at every resolution, and $4.1$--$4.5\times$ as much as FNO. Surprisingly, about half of its step is spent in pointwise ($1\times1$) convolutions and over a quarter in the bilinear \texttt{grid\_sample} of the warp; the spherical primitives themselves are comparatively cheap, with $12$--$14\%$ for the SHT pair and about $6\%$ for the Rodrigues map. The same two operations are the biggest factor for Flower2D, where strided convolutions take the place of the SHT pair.

\begin{table}[htbp]
  \centering
  \footnotesize
  \setlength{\tabcolsep}{3pt}
  \caption{Share of a training step (forward and backward) per component, on one H200 using each model's training batch size. ``vs FNO'' is the time per sample relative to FNO. Components are timed in isolation; the remainder (activations, reshapes, residual additions) is below $2\%$ of the step.}
  \label{tab:cost_breakdown}
  \begin{tabular}{@{}ll c ccccc c@{}}
    \toprule
    Grid & Model & vs FNO & Rodrigues & \texttt{grid\_sample} & SHT pair & strided conv & $1\times1$ conv & norm \\
    \midrule
    \multirow{2}{*}{$64\times128$}   & \textsc{Flower2D}  & $1.67\times$ & --     & $22\%$ & --     & $19\%$ & $50\%$ & $7\%$ \\
                                     & \textsc{\netname}  & $4.14\times$ & $6\%$  & $26\%$ & $14\%$ & --     & $48\%$ & $6\%$ \\
    \midrule
    \multirow{2}{*}{$128\times256$}  & \textsc{Flower2D}  & $1.74\times$ & --     & $25\%$ & --     & $19\%$ & $49\%$ & $7\%$ \\
                                     & \textsc{\netname}  & $4.36\times$ & $6\%$  & $28\%$ & $13\%$ & --     & $48\%$ & $6\%$ \\
    \midrule
    \multirow{2}{*}{$256\times512$}  & \textsc{Flower2D}  & $1.81\times$ & --     & $27\%$ & --     & $19\%$ & $48\%$ & $6\%$ \\
                                     & \textsc{\netname}  & $4.47\times$ & $6\%$  & $28\%$ & $12\%$ & --     & $48\%$ & $6\%$ \\
    \bottomrule
  \end{tabular}
\end{table}

\end{document}